\documentclass[journal]{IEEEtran}
\usepackage{amsmath,amsfonts}
\usepackage{algorithmic}
\usepackage{algorithm}
\usepackage{array}
\usepackage{textcomp}
\usepackage{stfloats}
\usepackage{url}
\usepackage{verbatim}
\usepackage{graphicx}
\usepackage{cite}
\usepackage[bookmarks=true,pdfencoding=auto,breaklinks,colorlinks,hypertexnames=false]{hyperref} %
\usepackage{amssymb}
\usepackage{xcolor}
\usepackage{dsfont}
\usepackage{siunitx}
\usepackage{colortbl}
\usepackage{booktabs}
\usepackage{multirow}
\usepackage{caption}
\usepackage{subcaption}
\usepackage{microtype}
\usepackage{cuted}
\usepackage{bibunits}
\usepackage{pifont}
\usepackage{flushend}

\usepackage{cleveref}

\let\citet\cite

\begin{document}

\definecolor{mygreen}{HTML}{00A64F}
\definecolor{myred}{HTML}{ED1B23}
\definecolor{color_best}{HTML}{C7E1CC}
\definecolor{color_second}{HTML}{E4EDBE}
\definecolor{color_third}{HTML}{FEFAC7}
\newcommand{\green}[1]{\textcolor{mygreen}{#1}}
\newcommand{\red}[1]{\textcolor{myred}{#1}}
\newcommand{\net}{OMEN}
\newcommand{\cmark}{\text{\ding{51}}}
\newcommand{\xmark}{\text{\ding{55}}}

\title{Class-Incremental Motion Forecasting}

\author{Anonymous Authors}
\author{Nicolas Schischka$^{1}$, Nikhil Gosala$^{1}$, B Ravi Kiran$^{2}$, Senthil Yogamani$^{3}$, and Abhinav Valada$^{1}$
\thanks{$^{1}$Department of Computer Science, University of Freiburg, Germany.}%
\thanks{$^{2}$Qualcomm SARL France.}%
\thanks{$^{3}$Automated Driving, Qualcomm Technologies, Inc.}%
}

\maketitle

\begin{abstract}
Motion forecasting enables autonomous vehicles to anticipate scene evolution by predicting the future trajectories of dynamic agents. However, existing approaches typically assume a closed-world setting with a fixed object taxonomy and access to high-quality perception, limiting their applicability in the real world where perception is imperfect, and new object classes may emerge over time. In this work, we introduce \textit{class-incremental motion forecasting}, a novel setting in which new object classes are sequentially introduced over time and future object trajectories are predicted directly from camera images. We propose the first end-to-end framework for this setting, which adapts to newly introduced classes while mitigating catastrophic forgetting of previously learned ones. Our method generates motion forecasting pseudo-labels for known classes and matches them with 2D instance masks from an open-vocabulary segmentation model. This 3D-to-2D keypoint voting mechanism filters inconsistent and overconfident predictions, while a query feature variance-based replay strategy samples informative past sequences to preserve prior knowledge. Extensive evaluations on nuScenes and Argoverse~2 show that our approach successfully preserves performance on known classes while effectively adapting to novel ones. We further demonstrate zero-shot transfer to real-world driving and show that the framework extends naturally to open- and closed-loop end-to-end class-incremental planning on nuScenes and NeuroNCAP. Code and models will be made publicly available at~\url{https://omen.cs.uni-freiburg.de}.%

\end{abstract}


\section{Introduction}
\label{sec:introduction}

\IEEEPARstart{A}{utonomous} vehicles operating in complex environments must be able to anticipate the future behavior of surrounding agents to ensure safe and efficient navigation. Motion forecasting addresses this need by predicting future trajectories of multiple agents in the scene, which, when combined with ego state, allows downstream planners to generate efficient and safe trajectories.
Existing approaches address the challenge of motion forecasting using sequence models such as LSTMs~\cite{cit:prediction_lstm-driver-intention1, cit:prediction_lstm-surr-vehicles, cit:prediction_lstm-cnn-socialpooling, cit:prediction_lstm-cnn-scichina}, graph neural networks~\cite{gao2020vectornet, liang2020learning}, and more recently, transformer-based architectures~\cite{cit:prediction_attn1, cit:prediction_attn2, wei2025parkdiffusion}. These approaches assume access to accurate, fully observed past trajectories derived from near-perfect perception systems. However, this assumption is rarely satisfied in the real world, with errors in detection and tracking cascading into forecasting and planning. End-to-end camera-based motion forecasting methods~\cite{cit:prediction_e2e-vip3d, hu2023planning, cit:prediction_e2e-sparsedrive} solve this limitation by predicting future trajectories directly from on-board multi-view images. By learning a differentiable mapping from raw sensor input to future object trajectories, these approaches reduce dependence on hand-crafted intermediate representations and improve robustness by circumventing the need for multiple decoupled modules.\looseness=-1

\IEEEpubidadjcol
Despite this progress, existing motion forecasting approaches are designed under a closed-world assumption, in which the set of agent categories is fixed in advance, and exhaustive annotations are available for every semantic class. However, this assumption often fails to generalize to the real world, where novel classes such as e-scooters may need to be routinely incorporated into the model. Under a static training paradigm, adding a new class requires re-annotating historical data with the new label, annotating newly recorded data with the old classes, and retraining the full model on the combined dataset, which is economically prohibitive and operationally impractical. Na\"ive fine-tuning on limited data containing only the new class typically leads to catastrophic forgetting~\cite{mccloskey1989catastrophic, ratcliff1990connectionist}, degrading forecasting performance on previously learned classes. These challenges are further exacerbated during real-world deployment, especially on edge devices, where storage and data retention constraints prevent model re-training on fully labeled datasets.

\begin{figure}[t]
    \centering
{\includegraphics[width=\linewidth]{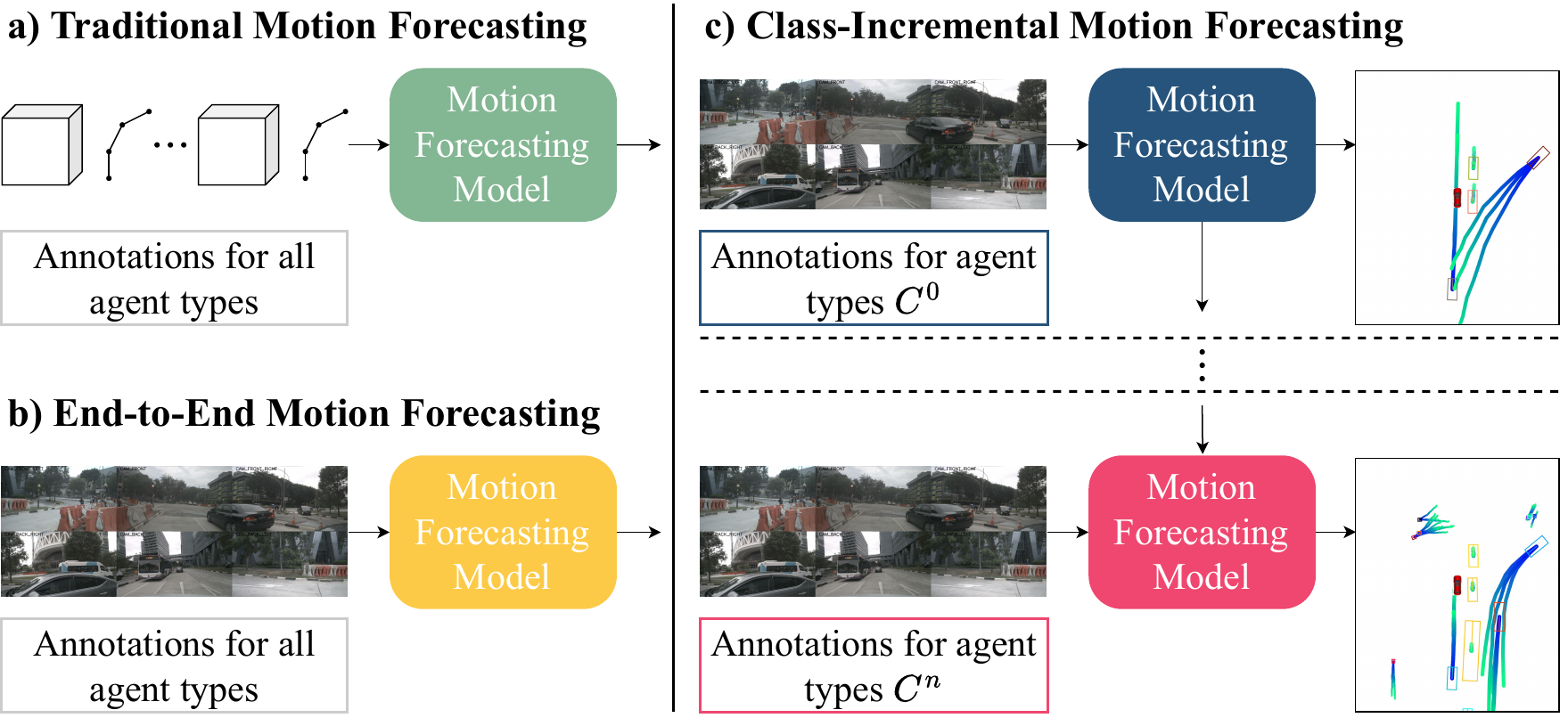}}
    \caption{Our approach is the first to tackle the problem of class-incremental motion forecasting. In contrast to (a) traditional and (b) end-to-end motion forecasting, (c) the underlying model is trained incrementally, with access to labels only for a subset of all classes $C^i$ and raw multi-view camera images. As a result, it continually learns to forecast the motion of all classes in an end-to-end manner, while handling imperfect object detections and successfully combating catastrophic forgetting.}
    \label{fig:teaser}
\end{figure}

In this work, we address these limitations by formalizing \textit{class-incremental motion forecasting} - an end-to-end incremental setting where models predict future trajectories directly from raw camera observations and incorporate new semantic classes from limited labeled data while retaining forecasting performance on previously learned ones without access to the original training set. 
To this end, we propose c\underline{O}ntinual \underline{M}otion Pr\underline{E}dictio\underline{N} (\net), the first approach specifically designed to address this novel setting. Our framework enables existing models to integrate new semantic classes using only a small set of annotated image sequences per class, while explicitly mitigating catastrophic forgetting. By eliminating the need for large-scale dataset re-annotation, full model re-training, and long-term storage of the entire dataset, our approach offers a scalable and memory-efficient solution for adding new classes to a motion forecasting model, as illustrated in \Cref{fig:teaser}.

\net~achieves this through two complementary mechanisms. First, we generate motion forecasting pseudo-labels for the sets of classes $\{C^0, \dots, C^{i-1}\}$ of the previous steps $\{0, \dots, i-1\}$ on the newly labeled dataset for a set of novel classes $C^i$ by leveraging the previously trained 3D detection head. This yields a unified labeled dataset covering all classes. To mitigate false positives and overly confident pseudo-label predictions, we employ 2D instance masks generated by a vision-language model (VLM), which can perform open-vocabulary segmentation, to filter out 3D detections that are inconsistent with the visual evidence via a 3D-to-2D keypoint voting mechanism. Second, we introduce a motion query-based experience replay strategy that prioritizes image sequences with a high sum of squared deviations in the latent motion-query feature space, corresponding to scenes with many moving objects. By sampling sequences with informative motion patterns, this strategy helps mitigate catastrophic forgetting while respecting memory constraints.

We perform extensive evaluations on the nuScenes~\cite{caesar2020nuscenes} and Argoverse 2~\cite{wilson2021argoverse} datasets and demonstrate that~\net~retains model performance and successfully mitigates catastrophic forgetting across multiple sequential class additions while continuing to adapt to new categories. Furthermore, we ablate the effectiveness of the individual network components and highlight that~\net~readily extends to class-incremental end-to-end open-loop planning on nuScenes and closed-loop planning on the NeuroNCAP benchmark~\cite{ljungbergh2024neuroncap}. Finally, we conduct real-world evaluations with our in-house self-driving perception vehicle and demonstrate \net's zero-shot capability.

Our key contributions can be summarized as follows:
\begin{enumerate}
    \item Formal introduction of the novel class-incremental motion forecasting task.
    \item \net, the first approach tailored for end-to-end class-incremental motion forecasting.
    \item A pseudo-labeling strategy with a VLM-guided error filtering mechanism via a 3D-to-2D keypoint voting to generate motion forecasting pseudo-labels.
    \item A variance-based experience replay selection mechanism based on the latent distribution of motion queries to alleviate long-term forgetting.
    \item Extensive experiments and a comprehensive ablation study on two real-world datasets.
    \item Demonstration of the natural extension of our model to class-incremental end-to-end planning as well as zero-shot capability for real-world deployment.
    \item Publicly available code upon acceptance at \url{https://omen.cs.uni-freiburg.de}.
\end{enumerate}

\section{Related Work}
\label{sec:related-work}

{\parskip=0pt
\noindent\textbf{Motion Forecasting} 
has been addressed using two primary paradigms, namely, \textit{tracklet-based forecasting} and \textit{end-to-end forecasting}. Tracklet-based forecasting approaches use pre-computed agent tracks as input and predict future trajectories. Given the sequential nature of such input data, recurrent networks have been extensively used to model future agent behavior. Several methods~\cite{cit:prediction_lstm-driver-intention1, cit:prediction_lstm-surr-vehicles, cit:prediction_lstm-intersection} leveraged LSTMs to model the behavior of multiple agents in the scene. LSTM-based networks have also been combined with CNNs to learn rich spatial information alongside temporal context~\cite{cit:prediction_lstm-cnn-socialpooling, cit:prediction_lstm-cnn-scichina}, and external scene information such as HD map data has also been introduced to better constrain motion predictions~\cite{cit:prediction_lstm-cnn-hdmap, wei2026parkdiffusion++}. However, LSTM networks suffer from vanishing and exploding gradients, which limit their temporal context. \citet{cit:prediction_lstm-shortcut} addressed this problem by introducing shortcut paths between LSTM blocks, while others replaced LSTM blocks with transformer-based attention to model long-range context~\cite{cit:prediction_attn1, cit:prediction_attn2, nayakanti2022wayformer, zhou2022hivt}. Several recent methods also enable the estimation of uncertainty in motion forecasting models~\cite{distelzweig2025stochasticity, marvi2025evidential, distelzweig2025motion}.

In contrast, end-to-end forecasting leverages a sequence of camera images to directly estimate the future states of all agents in the scene.
Traphic~\cite{cit:prediction_e2e-lstm-cnn-traphic} used an LSTM-CNN-based model to predict trajectories of road agents in dense videos, while~\citet{cit:prediction_e2e-attn1} proposed using a recurrent network with transformers for end-to-end forecasting. FIERY and Beverse~\cite{cit:prediction_e2e-fiery, cit:prediction_e2e-beverse} lifted multi-camera features to a spatio-temporal BEV representation for 3D perception tasks, including object forecasting. While PiP~\cite{cit:prediction_e2e-pip} and ViP3D~\cite{cit:prediction_e2e-vip3d} took additional HD maps as input and coupled them with transformer-based attention for end-to-end motion prediction, UniAD~\cite{hu2023planning} and SparseDrive~\cite{cit:prediction_e2e-sparsedrive} performed online map generation solely based on camera images.
Similar to the aforementioned approaches, \net~also follows an end-to-end paradigm and couples it with an evolving taxonomy of agent classes to perform class-incremental motion forecasting.}

{\parskip=2pt
\noindent\textbf{Class-Incremental Learning for Robot Perception}: 
The goal of class-incremental learning is to enable an existing model to progressively learn new classes over time without forgetting previously learned ones. Several approaches for image-based perception have been introduced, most of which used knowledge distillation and experience replay to prevent catastrophic forgetting~\cite{cit:ci_detection1, cit:ci_detection-mvcd, cit:ci_detection2, hindel2024taxonomy, vodisch2023covio}.
While \citet{cit:ci_detection-replay1} and \citet{cit:ci_detection-replay2} proposed the use of a replay buffer to tackle the forgetting problem for conventional detectors, CL-DETR~\cite{cit:ci_detection-cldetr} was the first approach to target incremental learning for DETR-like transformers by employing a calibration strategy to preserve the label distribution in the training set. COOLer~\cite{cit:ci_tracking-cooler} addressed class-incremental multi-object tracking by leveraging a contrastive loss formulation to better learn instance features of overlapping classes.

Only very few approaches have explored continual learning for tracklet-based object forecasting, in which new driving scenarios, rather than novel semantic classes, are sequentially added to the model. \citet{cit:ci_forecasting-grtp} proposed a divergence metric to determine dependencies between various driving scenes and uses a generative replay mechanism. CMP~\cite{cit:ci_forecasting-cmp} leveraged meta-representation learning to learn a sparse scene representation and coupled it with a replay buffer where samples are chosen based on their representation similarity. Instead of producing a unified model suitable for all scenarios, DECODE~\cite{cit:ci_forecasting-decode} incrementally generated specialized models for distinct domains. This is achieved by using a hypernetwork to generate specialized network parameters and by leveraging normalizing flow to select the best model during inference. %
To the best of our knowledge, no approach has been proposed thus far for either continual or class-incremental end-to-end forecasting, which we address in this work.}

\begin{figure*}[t]
    \centering
    {\includegraphics[width=0.95\linewidth]{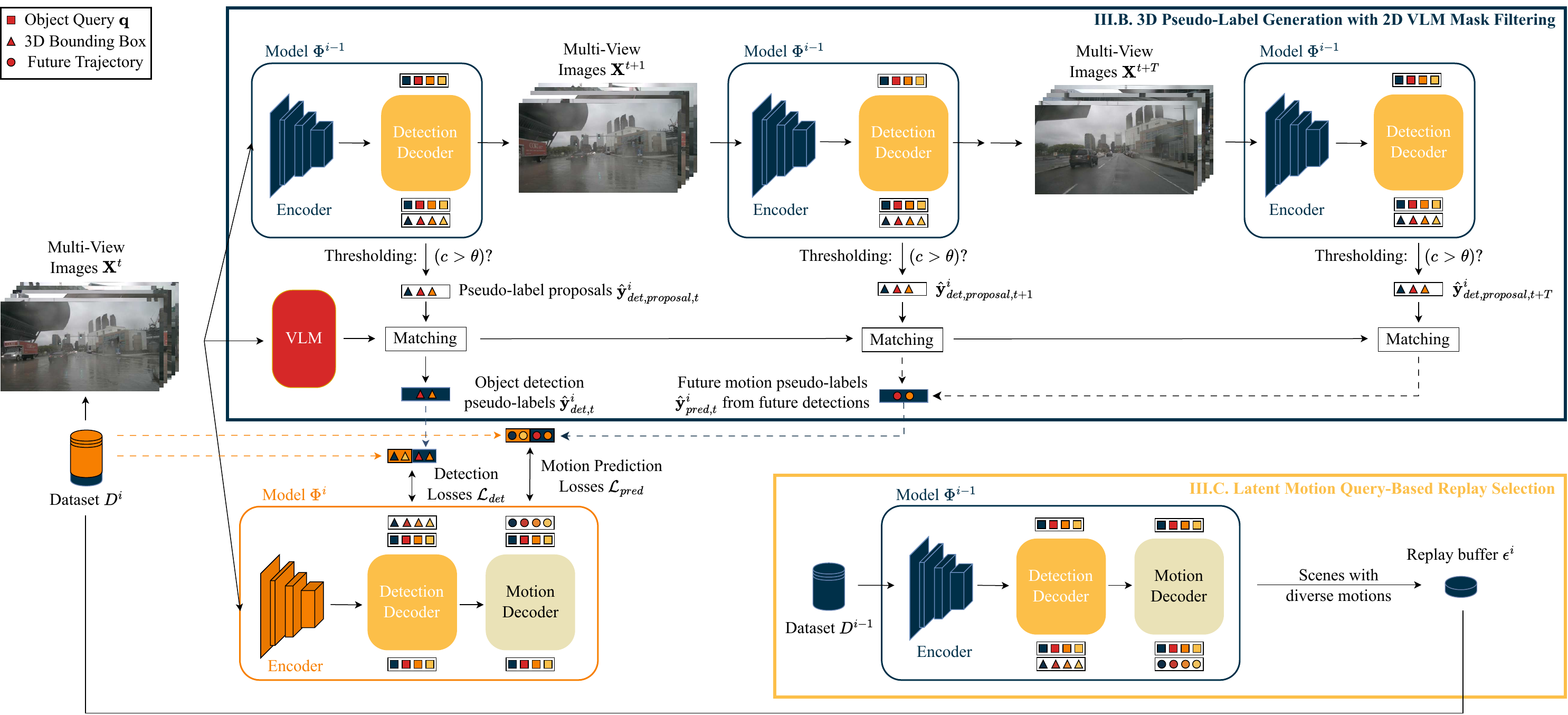}}
    \caption{Illustration of the proposed \net~architecture. At each incremental step $i$, we create detection and motion forecasting pseudo-labels for the old categories with the old model $\Phi^{i-1}$, filter them in an offline process via a 3D-to-2D keypoint voting mechanism with the 2D instance masks output by a VLM, and add them to the detection ($\triangle$) and motion forecasting ($\circ$) ground truth of $D^i$, as detailed in \Cref{subsec:pseudo-labels}. Furthermore, a replay buffer is created as described in \Cref{subsec:replay}, based on the latent motion query space of the old model.}
    \label{fig:network-architecture}
\end{figure*}

\section{Technical Approach}
\label{sec:technical-approach}
In this section, we formally define the class-incremental motion prediction problem and present \net, our novel end-to-end class-incremental framework for camera-based motion forecasting, illustrated in~\Cref{fig:network-architecture}. Specifically, we outline our offline 3D pseudo-label generation module with 2D VLM mask filtering to produce accurate motion forecasting pseudo-labels in \Cref{subsec:pseudo-labels}, and present our novel latent motion query-based experience replay strategy to address long-term forgetting in \Cref{subsec:replay}.

\subsection{Class-Incremental Motion Prediction}
\label{subsec:task}
We define class-incremental motion forecasting as the task of predicting the future trajectories of all agents in a scene directly from camera images $\mathbf{X}$ and an evolving set of semantic classes $C^i$ over time.
Analogously to class-incremental object detection~\cite{shmelkov2017incremental}, the model is trained in incremental steps, where it has access to only a disjoint set of labels. In other words, after step $i-1$, the model has to retain knowledge of previously learned classes without accessing their ground truth samples in all subsequent steps $[i, i+1, ...)$.
To facilitate this, we split the underlying dataset into several subsets, each with a distinct set of class labels.
Thus, the dataset $D^i$ in the incremental step $i$ consists of $N_S$ sequences, each with multiple sequential samples. Each sample contains $M$ multi-view RGB images $\mathbf{X}$ in addition to the 3D bounding box parameters $\{x, y, z, w, l, h, yaw, v_x, v_y, v_z\}$, class labels, and future agent trajectories $\boldsymbol{\tau}_{t:t+T} = \{( (x, y)_{t+1}, (x, y)_{t+2}, \ldots, (x, y)_{t+T} )\}
$ for all objects of the set of classes $C^i$.
Note that the images may still contain objects from past or future classes, but the ground truth includes annotations for only the subset of classes $C^i$. 
An exception from this formulation is a small exemplar replay buffer $\epsilon^i$ that contains a fixed size of $N_\text{replay} \ll N_S$ sequences from the previous datasets $\{D^0, \dots, D^{i-1}\}$ with annotations for classes $C^0, \dots, C^{i-1}$ of the previous steps.

\subsection{3D Pseudo-Label Generation with 2D VLM Mask Filtering}\label{subsec:pseudo-labels}
\subsubsection{Pseudo-Labeling for Motion Prediction}
In each incremental step $i$, we use the model of the previous step $\Phi^{i-1}$ in an offline process to generate 3D object detection pseudo-labels $\hat{\mathbf{y}}^i_{det}$ for all classes observed in previous steps $\{0, \dots, i-1\}$ with the $M$ multi-view images $\mathbf{X}_i$ as input. As the framework is designed for end-to-end DETR-based architectures, the object detection output $\hat{\mathbf{y}}^{i-1}_{det}$ consists of a set of $N$ predictions, each with 3D bounding box parameters and confidence score $c$.
Following incremental 2D detection approaches, we retain only predictions whose confidence $c$ exceeds a threshold $\theta$, and denote these as pseudo-label proposals $\hat{\mathbf{y}}^i_{det, proposal}$.

A straightforward idea is to proceed in a similar manner to retrieve motion prediction pseudo-labels $\hat{\mathbf{y}}^i_{motion, proposal} \in \mathds{R}^{\hat{N} \times T \times 2}$ for the next $T$ time steps for all $\hat{N}$ objects that pass the threshold $\theta$ with the old model $\Phi^{i-1}$.
However, we observe that a more accurate way to capture the motion of moving objects, especially with non-linear trajectories, is to use the 3D object positions estimated by the detection decoder for the images from future time steps up to $T$.
Since the underlying model inherently performs end-to-end tracking via query propagation and each query has a unique identity, we do not require any external method to derive object associations and, in turn, future positions across different timestamps.
Formally, we compose the motion forecasting pseudo-label $\hat{\mathbf{y}}^i_{j, motion, proposal}$ for object $j$ as [$\tilde{\mathbf{y}}_{j, t+1}, \dots, \tilde{\mathbf{y}}_{j, t+T}$].
Here, we transform the center of the object $\{x, y, z\}_{\tau}$ at each future step $\tau$ by the homogeneous transformation matrix $\mathbf{T}_{\tau \rightarrow t}$ from the future ego-vehicle frame to the current ego-coordinate system:
\begin{equation}
    \tilde{\mathbf{y}}^i_{j, motion, proposal, t+\tau} = 
\mathbf{T}_{\tau \rightarrow t}
\begin{bmatrix} x \\ y \\ z \\ 1 \end{bmatrix}_{\tau}.
\end{equation}

\subsubsection{VLM-Guided Pseudo-Label Filtering via 3D-to-2D Keypoint Voting}
We observe that the model's confidence $c$ increases over steps, leading to a higher rate of false positives (FPs).
Since we are using a model that operates directly on raw sensor input, it is crucial to mitigate this effect, as it can learn and accumulate motion patterns and object instances that do not exist in the actual scene.
Moreover, improving upstream object detection and tracking performance indirectly influences end-to-end motion prediction results by yielding more accurate motion-forecasting pseudo-labels in subsequent steps.
We employ a VLM, Grounded SAM~2~\cite{ren2024grounded}, capable of open-vocabulary instance segmentation to reduce the number of FPs via a novel 3D-to-2D filtering mechanism.
By doing so, the model maintains consistent confidence over time and remains calibrated.
While utilizing foundation models can help guide the annotation process of old classes, their zero-shot generalization may not extend to newly emerging classes in the real world without further adaptation or fine-tuning. Therefore, we only use the VLM for old classes.

\begin{figure}[t]
    \centering
    {\includegraphics[width=\linewidth]{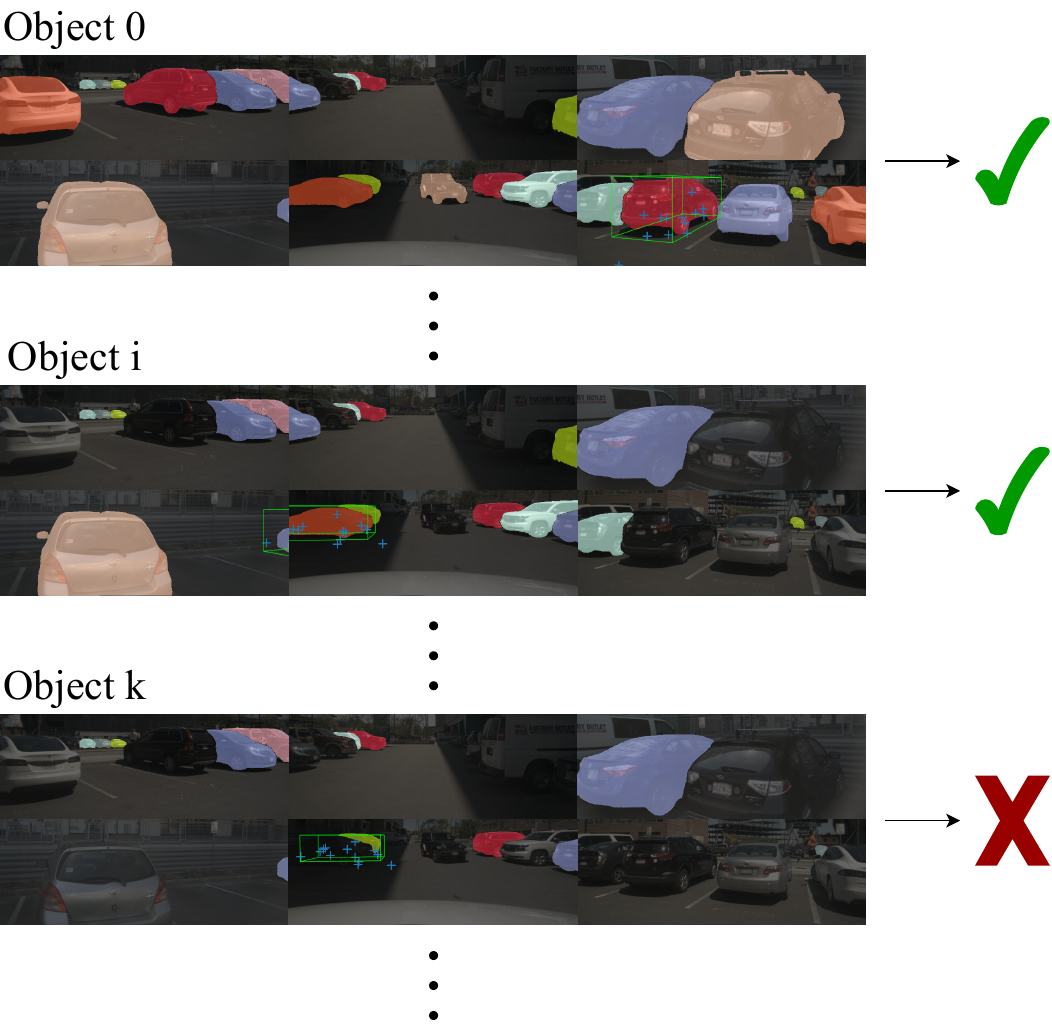}}
    \caption{Visualization of the VLM-guided pseudo-label filtering for objects of the \textit{car} class. Whenever a bounding box is associated with a VLM instance mask, the mask is removed for the remaining objects. Therefore, object $k$ is not added to the set of pseudo-labels since it refers to the same object as object $i$ that has a higher confidence. Note that objects $i$ and $k$ do not overlap in 3D space, which means they cannot be filtered based on 3D IoU.}
    \label{fig:mask_filtering}
\end{figure}

First, we integrate the foundation model by prompting it with the classes from previous steps, separated by periods, e.g., ``car. pedestrian.'', and using its predicted 2D instance masks within the 2D object bounding boxes of Grounding DINO~\cite{liu2024grounding}.
To associate our 3D object detections with these 2D masks, we then use the keypoint feature sampling generator of the underlying sparse 3D object detection method.
Given the anchor box for an object and its feature query, this generator returns multiple fixed and learnable 3D points, which are usually used for feature sampling during regular model training.
In the pseudo-label filtering step, however, we leverage these keypoints for a voting system: We project each of the $N_K$ keypoints $\mathbf{K}_{i}$ to all of the $v \in \mathcal{V}$ multi-view image planes with the projection function $\pi_v(\cdot)$.
If the absolute majority of the projected points for an object of class $c$ lies in at least one view $v$ inside an instance mask $\mathcal{M}_v^{c}$ of the same class, we keep this pseudo-label proposal as a true positive pseudo-label $\hat{\mathbf{y}}_{j} = \{\hat{\mathbf{y}}_{j, det}, \hat{\mathbf{y}}_{j, motion}\}$ and add it to the set of ground truth labels $\mathbf{y}$.
Formally, this condition of the voting of keypoints is defined as follows:
\begin{equation}
\exists v \in \mathcal{V}, \;\; \exists M \in \mathcal{M}_v^{c} \;:\; 
\frac{1}{N_K}
\sum_{i=1}^{N_K}
\mathds{I}\!\left(
\pi_v(\mathbf{K}_i) \in M
\right)
> \tfrac{1}{2}
\end{equation}
Simultaneously, we remove the associated 2D bounding box and mask from the set of masks $\mathcal{M}_v^{c}$.
As a result, we do not match any remaining proposal with the same mask again, as visualized for the class \textit{car} in~\Cref{fig:mask_filtering}.
As object $k$ and object $i$ refer to the same object predicted by the VLM, this mechanism removes object $k$, which has lower confidence. Importantly, the bounding boxes of $i$ and $k$ have a 3D Intersection over Union (IoU) of 0, i.e., we cannot apply a filtering based solely on overlap.
Note that if we perceive an object in multiple cameras, we remove the corresponding masks from all views.
By traversing the object list in descending order, we ensure that only the most confident object is matched to an object predicted by the VLM.

In cases where the VLM misses an object, this process would result in the removal of a true positive.
Therefore, we still add unmatched pseudo-label proposals to the set of ground truth labels and include them in the bipartite matching during training, but set their weights to zero in all loss calculations. %
Additionally, we do not apply filtering to objects that have already passed the confidence threshold $\theta$ in a previous frame, since FP predictions typically occur only for the first observation of an object, and the applied VLM outputs instance masks only for visible objects, which would filter out temporarily occluded objects.

\begin{table*}[h]
\scriptsize
\centering
\caption{Class-incremental motion prediction results on the nuScenes validation set for the proposed per-class incremental setting. We report all metrics for the first class car (1), the second class pedestrian (2), and all classes (All). $\dagger$ means that for this experiment, our proposed pseudo-labeling from the future detections was used. The best results are marked \textbf{bold}, and the second-best \underline{underlined}.\label{tab:quant_results_per_class}}
\begin{tabular}{l|cc>{\columncolor[gray]{0.9}}c|cc>{\columncolor[gray]{0.9}}c|cc>{\columncolor[gray]{0.9}}c|cc>{\columncolor[gray]{0.9}}c|>{\columncolor[gray]{0.9}}c>{\columncolor[gray]{0.9}}c>{\columncolor[gray]{0.9}}c} 
    \toprule
    \multirow{2}{*}{Method}
    & \multicolumn{3}{c|}{EPA (\%)$\uparrow$}
    & \multicolumn{3}{c|}{AP\_f$_{static}$ (\%)$\uparrow$}
    & \multicolumn{3}{c|}{AP\_f$_{linear}$ (\%)$\uparrow$}
    & \multicolumn{3}{c|}{AP\_f$_{non-linear}$ (\%)$\uparrow$}
    & \multicolumn{3}{>{\columncolor[gray]{0.9}}c}{mAP\_f (\%)$\uparrow$}\\
    \cmidrule{2-16}
    & (1) & (2) & All
    & (1) & (2) & All
    & (1) & (2) & All
    & (1) & (2) & All
    & (1) & (2) & All\\
    \midrule
    Joint Training & 45.92 & 35.49 & 30.26 & 54.27 & 31.99 & 34.23 & 29.01 & 31.10 & 18.78 & 14.97 & 14.11 & 6.60 & 32.75 & 25.73 & 19.87\\
    \midrule\midrule
    Forgetting & 0.0 & 0.0 & 2.31 & 0.0 & 0.0 & 3.39 & 0.0 & 0.0 & 1.28 & 0.0 & 0.0 & 0.64 & 0.0 & 0.0 & 1.77\\
    LwF~\cite{li2017learning} & 25.65 & 6.07 & 13.05 & 33.26 & 14.35 & 18.40 & 16.80 & 7.88 & 9.23 & 6.89 & 2.61 & 2.71 & 18.98 & 8.28 & 10.11\\
    Pseudo-Labeling & 3.25 & -1.38 & 8.47 & 41.01 & 13.51 & 22.05 & 10.61 & \textbf{19.73} & 10.62 & 2.94 & \underline{8.05} & 2.73 & 18.19 & 13.76 & 11.80\\
    Pseudo-Labeling$\dagger$ & 19.82 & 6.87 & 12.12 & 46.13 & 17.76 & 22.78 & \underline{25.06} & 18.40 & 14.47 & 10.63 & 7.35 & 3.93 & 27.27 & 14.50 & 13.73\\
    CL-DETR$\dagger$~\cite{cit:ci_detection-cldetr} & \textbf{42.93} & \underline{20.64} & 19.65 & \textbf{49.99} & \textbf{20.24} & \underline{24.08} & 24.58 & 16.27 & \underline{14.90} & \underline{10.96} & 7.70 & \underline{4.09} & \underline{28.51} & \underline{14.74} & \underline{14.35}\\
    \midrule
    Ours & \underline{39.34} & \textbf{20.80} & \textbf{21.95} & \underline{47.95} & \underline{19.86} & \textbf{26.01} & \textbf{28.86} & \underline{19.50} & \textbf{16.24} & \textbf{13.04} & \textbf{8.10} & \textbf{4.57} & \textbf{29.95} & \textbf{15.82} & \textbf{15.60}\\
    \bottomrule
\end{tabular}
\end{table*}

\subsection{Latent Motion Query-Based Replay Selection}\label{subsec:replay}
Despite the aforementioned strategies to improve pseudo-label quality, we observe degradation across incremental steps.
Moreover, if old classes were completely absent in an intermediate step, pseudo-labeling would not be suitable on its own.
Therefore, we create a small replay buffer $\epsilon^i$ that contains $N_{replay}$ sequences $\mathbf{S}$. It is important to emphasize that we should employ a strategy for selecting sequences rather than single samples, since motion forecasting models need context from previous time steps.
More specifically, we keep the number of sequences $N_{replay}$ in the buffer constant and divide it evenly among all $\mathcal{C} = \{C_0, \dots, C_{i-1}\}$ previously learned classes.
This results in the following definition of the buffer of step $i$ with $P = N_{replay} / \lvert \mathcal{C} \rvert$:
\begin{equation}
    \epsilon^i = \{\mathbf{S}_0^{C_0}, \dots, \mathbf{S}_P^{C_0}, \dots, \mathbf{S}_0^{C_{i-1}}, \dots, \mathbf{S}_P^{C_{i-1}}\}.
\end{equation}

Selecting the set of past sequences is non-trivial, as each sequence contains multiple samples that vary in the number of objects and object motions, with many being static.
As a result, it may be suboptimal to use a purely image-feature-based approach as it does not account for different types of trajectories.
The latter also holds true for strategies that incorporate statistics about dataset distributions, such as~\cite{cit:ci_detection-cldetr}.
In contrast, by taking advantage of the latent representation of size $D$ of the end-to-end model, we base the replay buffer on statistics about the motion queries $\mathbf{Q}_{motion} \in \mathds{R}^{N \times D}$ corresponding to all ground truth objects in each sample.
To retrieve a motion query $\mathbf{q}_{j, motion}$ for each object instance $j$ in the dataset $D^{i-1}$, we provide all samples of all sequences from the previous step $i-1$ to the model.
Then, we adapt the Hungarian matching from training to associate each ground truth instance with a model prediction.
Subsequently, we calculate the mean motion query $\mathbf{\bar{q}}_{c}$ of all $N_{total}$ objects in $D^{i-1}$ per class $c$:
\begin{equation}
    \mathbf{\bar{q}}_{c} = \frac{\sum_{j=1}^{N_{total}} \mathbf{q}_j \, \mathds{1}_{\{y_j = c\}}}{\sum_{j=1}^{N_{total}} \mathds{1}_{\{y_j = c\}}}.
\end{equation}
The final score $s_{k, c}$ for class $c$ in sequence $k$ is then defined as the sum of squared deviations from this mean query over all $N_{inst, k}$ object instances:
\begin{equation}
    s_{k, c} = \sum_j^{N_{inst, k}} (\mathbf{q}_j - \mathbf{\bar{q}}_{c})^2 \mathds{1}_{\{y_j = c\}}.
\end{equation}
Finally, we select the $P$ sequences with the highest scores as replay samples per class because they generally represent scenes with a moderate to high number of \textit{moving} objects with linear and non-linear future trajectories.

\subsection{Extension to Class-Incremental Planning}
As the proposed framework for class-incremental motion prediction inherently leads to a better understanding of the surrounding environment, it should also directly positively influence the planning of the ego-vehicle trajectory $\boldsymbol{\tau}_{ego, t:t+T} = \{( (x, y)_{ego, t+1}, (x, y)_{ego, t+2}, \ldots, (x, y)_{ego,t+T} \})$.
Therefore, to achieve class-incremental planning, we concatenate a query for the ego-vehicle $\mathbf{q}_{ego} \in \mathds{R}^{1 \times D}$ to the set of $N$ $D$-dimensional object queries $\mathbf{Q} \in \mathds{R}^{N \times D}$:
\begin{equation}
    \tilde{\mathbf{Q}} = concatenate(\mathbf{Q}, \mathbf{q}_{ego}),
\end{equation}
and additionally predict its future motion under the setup of only having access to a set of class labels $C^i$ at a time, following~\cite{cit:prediction_e2e-sparsedrive}.
\section{Experimental Results}
\label{sec:experiments}

\begin{table}[t]
\scriptsize
\setlength{\tabcolsep}{8.5pt}
\centering
\caption{Class-incremental 3D object detection and tracking results on the nuScenes validation set for the proposed per-class incremental setting. We report all metrics for the first class car (1) and all classes (All).
$\dagger$ means that for this experiment, our proposed pseudo-labeling from the future detections was used. The best results are marked \textbf{bold}, and the second-best \underline{underlined}.\label{tab:quant_results_det}}
\begin{tabular}{l|c>{\columncolor[gray]{0.9}}c|>{\columncolor[gray]{0.9}}c>{\columncolor[gray]{0.9}}c||>{\columncolor[gray]{0.9}}c>{\columncolor[gray]{0.9}}c} 
    \toprule
    \multirow{2.5}{*}{Method}
    & \multicolumn{2}{c|}{mAP (\%)$\uparrow$} 
    & \multicolumn{2}{>{\columncolor[gray]{0.9}}c||}{NDS $(\%)\uparrow$} 
    & \multicolumn{2}{>{\columncolor[gray]{0.9}}c}{AMOTA $(\%)\uparrow$}\\
    \cmidrule{2-7}
    & (1) & All
    & (1) & All
    & (1) & All\\
    \midrule
    Joint Training & 59.6 & 37.1 & 68.1 & 48.2 & 49.9 & 30.7\\
    \midrule\midrule
    Forgetting & 0.0 & 4.6 & 0.0 & 6.1 & 0.0 & 5.2\\
    LwF~\cite{li2017learning} & 39.1 & 20.8 & 56.0 & 37.9 & 24.5 & 13.6\\
    Pseudo-Labeling & 45.9 & 25.6 & 58.1 & 41.0 & 29.2 & 19.4\\ %
    Pseudo-Labeling$\dagger$ & 49.9 & 25.8 & 61.7 & 40.8 & 41.3 & 20.8\\
    CL-DETR$\dagger$~\cite{cit:ci_detection-cldetr} & \textbf{53.5} & \underline{27.5} & \textbf{64.5} & \underline{42.0} & \underline{52.1} & \underline{25.2}\\
    \midrule
    Ours & \underline{52.3} & \textbf{29.2} & \underline{64.4} & \textbf{43.8} & \textbf{52.7} & \textbf{28.0}\\
    \bottomrule
\end{tabular}
\end{table}

\begin{table*}
\scriptsize
\centering
\caption{Class-incremental motion prediction results on the nuScenes validation set for the proposed group-incremental setting, which consists of a base training with four classes (1-4), an incremental step with one class (5), followed by a second incremental step with two classes (6-7). We report all metrics for the individual steps and all classes (All). $\dagger$ means that for this experiment, our proposed pseudo-labeling from the future detections was used. The best results are marked \textbf{bold}, and the second-best \underline{underlined}.\label{tab:quant_results}}
\begin{tabular}{l|ccc>{\columncolor[gray]{0.9}}c|ccc>{\columncolor[gray]{0.9}}c|ccc>{\columncolor[gray]{0.9}}c| >{\columncolor[gray]{0.9}}c >{\columncolor[gray]{0.9}}c >{\columncolor[gray]{0.9}}c >{\columncolor[gray]{0.9}}c} 
    \toprule
    \multirow{2.5}{*}{Method}
    & \multicolumn{4}{c|}{AP\_f$_{static}$ (\%)$\uparrow$} 
    & \multicolumn{4}{c|}{AP\_f$_{linear}$ (\%)$\uparrow$} 
    & \multicolumn{4}{c|}{AP\_f$_{non-linear}$ (\%)$\uparrow$}
    & \multicolumn{4}{>{\columncolor[gray]{0.9}}c}{mAP\_f (\%)$\uparrow$}\\
    \cmidrule{2-17}
    & (1-4) & (5) & (6-7) & All
    & (1-4) & (5) & (6-7) & All
    & (1-4) & (5) & (6-7) & All
    & (1-4) & (5) & (6-7) & All\\
    \midrule
    Joint Training & 30.61 &
    31.99 & 42.60 & 34.23 & 15.07 & 31.10 & 20.04 & 18.78 & 7.85 & 14.11 & 0.36 & 6.60 & 17.85 & 25.73 & 21.00 & 19.87\\
    \midrule\midrule
    Forgetting & 0.0 & 0.0 & \underline{28.42} & 8.12 & 0.0 & 0.0 & \underline{13.49} & 3.85 & 0.0 & 0.0 & 0.10 & 0.03 & 0.0 & 0.0 & 14.00 & 4.00\\
    LwF~\cite{li2017learning} & 15.61 & 17.63 & 27.71 & 19.36 & 3.88 & 8.58 & 9.16 & 6.06 & 1.59 & 4.31 & 0.18 & 1.58 & 7.03 & 10.17 & 12.35 & 9.00\\
    Pseudo-Labeling & 19.92 & 15.36 & 26.01 & 21.01 & 4.22 & 19.04 & 3.74 & 6.20 & 1.89 & \textbf{8.44} & 0.13 & 2.32 & 8.68 & 14.28 & 9.96 & 9.84\\
    Pseudo-Labeling$\dagger$ & 20.53 & 18.90 & 23.64 & 21.18 & 6.07 & \underline{19.67} & 8.98 & 8.84 & 2.21 & \underline{8.23} & \textbf{0.40} & \textbf{2.55} & 9.60 & \underline{15.60} & 11.00 & 10.86\\
    CL-DETR$\dagger$~\cite{cit:ci_detection-cldetr} & \underline{22.00} & \underline{20.49} & 27.68 & \underline{23.41} & \underline{6.59} & 16.74 & \textbf{14.88} & \underline{10.41} & \underline{2.31} & 6.23 & 0.11 & 2.24 & \underline{10.30} & 14.49 & \underline{14.23} & \underline{12.02}\\
    \midrule
    Ours & \textbf{22.74} & \textbf{20.59} & \textbf{30.78} & \textbf{24.73} & \textbf{7.18} & \textbf{20.75} & 12.69 & \textbf{10.69} & \textbf{2.75} & 6.44 & \underline{0.17} & \underline{2.54} & \textbf{10.89} & \textbf{15.93} & \textbf{14.55} & \textbf{12.65}\\
    \bottomrule
\end{tabular}
\end{table*}

\begin{table*}
\scriptsize
\centering
\caption{Class-incremental 3D object detection and motion prediction results on the Argoverse 2 validation set for the proposed incremental 20-2 setting with a total of four steps. We set the evaluation range to 50 $m$, following the AV2 End-to-End Forecasting settings for motion prediction. $\dagger$ indicates that, for this experiment, we used our proposed pseudo-labeling based on future detections. The best results are marked \textbf{bold}, and the second-best \underline{underlined}.\label{tab:quant_results_av2}}
\begin{tabular}{l|cc>{\columncolor[gray]{0.9}}c||cc>{\columncolor[gray]{0.9}}c|cc>{\columncolor[gray]{0.9}}c|cc>{\columncolor[gray]{0.9}}c| >{\columncolor[gray]{0.9}}c >{\columncolor[gray]{0.9}}c >{\columncolor[gray]{0.9}}c >{\columncolor[gray]{0.9}}c} 
    \toprule
    \multirow{2.5}{*}{Method}
    & \multicolumn{3}{c||}{CDS (\%)$\uparrow$}
    & \multicolumn{3}{c|}{AP\_f$_{static}$ (\%)$\uparrow$} 
    & \multicolumn{3}{c|}{AP\_f$_{linear}$ (\%)$\uparrow$} 
    & \multicolumn{3}{c|}{AP\_f$_{non-linear}$ (\%)$\uparrow$}
    & \multicolumn{3}{>{\columncolor[gray]{0.9}}c}{mAP\_f (\%)$\uparrow$}\\
    \cmidrule{2-16}
    & (1-20) & (21-26) & All
    & (1-20) & (21-26) & All
    & (1-20) & (21-26) & All
    & (1-20) & (21-26) & All
    & (1-20) & (21-26) & All\\
    \midrule
    Joint Training & 31.21 & 16.74 & 27.87 & 44.74 & 26.67 & 41.13 & 27.18 & 16.29 & 25.00 & 18.55 & 29.41 & 21.45 & 32.63 & 24.32 & 30.83\\
    \midrule\midrule
    Forgetting & 0.0 & 0.72 & 0.17 & 0.0 & 1.74 & 0.35 & 0.0 & 0.09 & 0.02 & 0.0 & 2.49 & 0.66 & 0.0 & 1.46 & 0.32\\
    LwF~\cite{li2017learning} & 7.73 & 11.75 & 8.66 & 10.85 & \underline{21.08} & 12.89 & 7.88 & 9.79 & 8.26 & 6.90 & 14.86 & 9.02 & 8.91 & 15.69 & 10.38\\
    Pseudo-Labeling & \underline{29.01} & \underline{15.14} & \underline{25.81} & \textbf{41.43} & 20.62 & \underline{37.27} & 25.43 & 11.83 & 22.71 & 11.87 & \textbf{27.38} & 16.00 & 29.07 & \underline{19.99} & 27.10\\
    Pseudo-Labeling$\dagger$ & 28.54 & \textbf{15.27} & 25.48 & 40.91 & 19.76 & 36.68 & \underline{26.89} & 12.96 & \underline{24.10} & \underline{15.65} & \underline{26.25} & \underline{18.47} & \underline{30.22} & 19.66 & \underline{27.94}\\
    CL-DETR$\dagger$~\cite{cit:ci_detection-cldetr} & 28.43 & 14.46 & 25.20 & 40.30 & 18.59 & 35.96 & 25.06 & \textbf{14.38} & 22.93 & 13.75 & 20.49 & 15.55 & 28.90 & 17.88 & 26.51\\
    \midrule
    Ours & \textbf{29.19} & 14.56 & \textbf{25.82} & \underline{41.37} & \textbf{22.27} & \textbf{37.55} & \textbf{28.97} & \underline{14.18} & \textbf{26.01} & \textbf{18.48} & 24.74 & \textbf{20.15} & \textbf{31.79} & \textbf{20.54} & \textbf{29.36}\\
    \bottomrule
\end{tabular}
\end{table*}

\begin{table}
\scriptsize
\centering
\caption{Class-incremental forgetting percentage points (FPP) regarding mAP\_f for the first set of classes $C^0$ on the three proposed splits. Negative values indicate improvement over the base training, positive values indicate forgetting. $\dagger$ means that for this experiment, we used our proposed pseudo-labeling from the future detections. The best results are marked \textbf{bold}, and the second-best \underline{underlined}.\label{tab:forgetting_metrics}}
\begin{tabular}{l|cc|c}
    \toprule
    \multirow{2.5}{*}{Method} & \multicolumn{2}{c|}{nuScenes} & \multirow{3}{*}{Argoverse 2}\\
    \cmidrule(lr){2-3}
    & 7 steps & 3 steps & \\
    \midrule
    Forgetting & 28.92 & 10.15 & 32.93 \\
    LwF~\cite{li2017learning} & 9.94 & 3.12 & 24.02\\
    Pseudo-Labeling & 10.73 & 1.47 & 3.86 \\
    Pseudo-Labeling$\dagger$ & 1.65 & 0.55 & \underline{2.71} \\
    CL-DETR$\dagger$~\cite{cit:ci_detection-cldetr} & \underline{0.41} & \underline{$-$0.15} & 4.03 \\
    \midrule
    Ours & \textbf{$-$1.03} & \textbf{$-$0.74} & \textbf{1.14} \\
    \bottomrule
\end{tabular}
\end{table}

\begin{figure*}[t]
    \centering
    {\includegraphics[width=\linewidth]{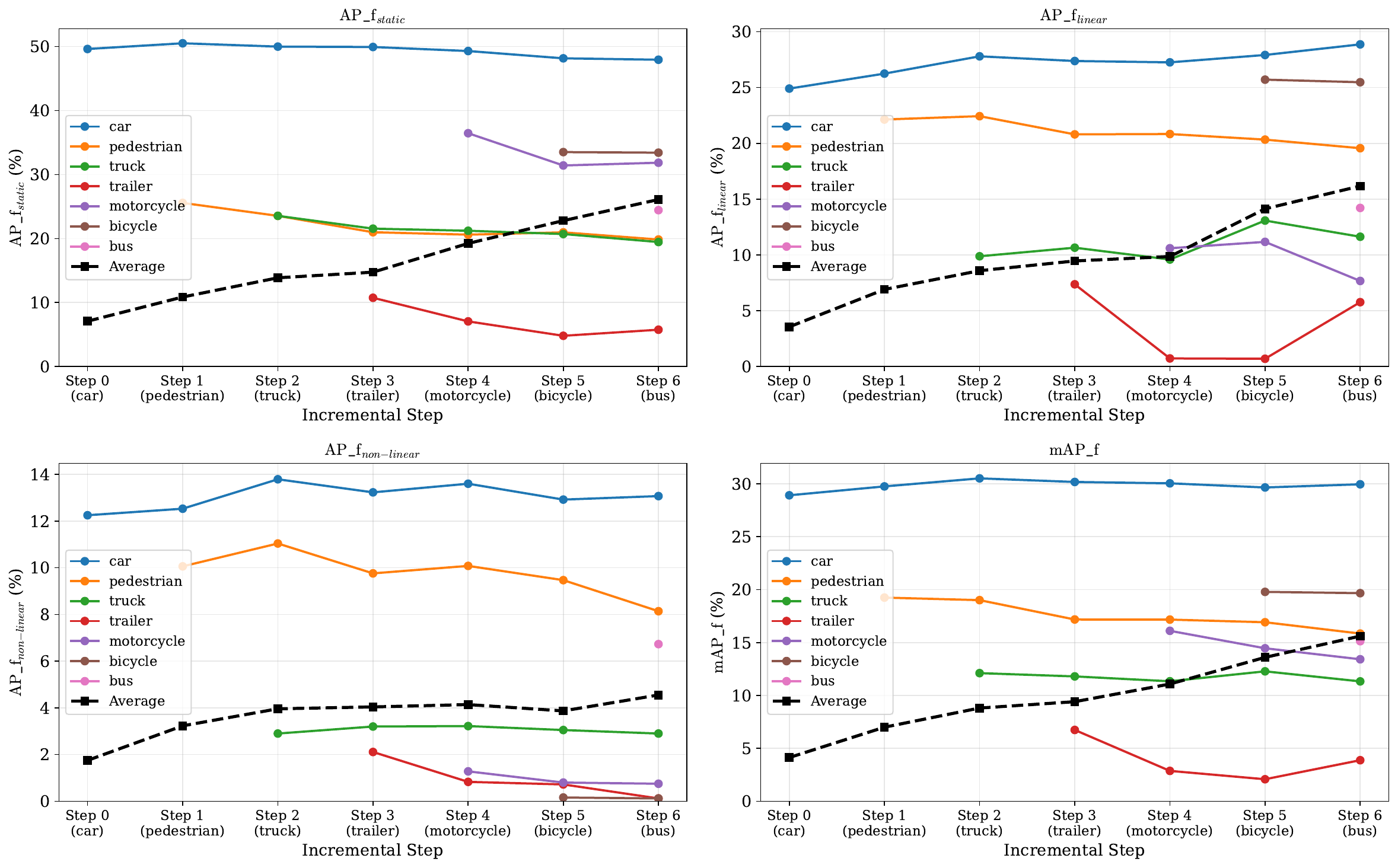}}
    \caption{Class-incremental motion prediction results per class and per incremental step on the nuScenes validation set for the proposed per-class incremental setting. While the class-averaged forecasting metrics increase over incremental steps, it can be observed that classes introduced at later steps, representing rarer classes, are more prone to forgetting. Moreover, some classes experience an increase in forecasting performance for intermediate steps due to the coupling of agents from different classes.}
    \label{fig:results_per_class}
\end{figure*}

We design our experimental setup to answer the following:
\begin{enumerate}
    \item Does our proposed framework effectively mitigate forgetting and achieve similar accuracy as a theoretical upper bound training with all labels?
    \item Does our approach generalize to real-world autonomous driving in a zero-shot setting?
    \item Can our approach be naturally extended to support class-incremental planning?
\end{enumerate}
    
\subsection{Datasets}
{\parskip=2pt
\noindent\textbf{nuScenes}}:
To set up two class-incremental motion forecasting scenarios, we use the large-scale nuScenes dataset~\cite{caesar2020nuscenes}, captured in Singapore and Boston.
For all approaches, we resize the six multi-view RGB images captured at $2$~Hz to $256 \times 704$ pixels as model inputs and consider only classes that may contain moving objects.

We create two distinct training data splits, each presenting its own specific challenges: a per-class incremental and a group-incremental split.
In the per-class incremental setting, each step contains annotations for only a single class, resulting in six incremental steps that are particularly prone to forgetting.
To train the underlying end-to-end model reasonably well, we randomly sample $300$ sequences per class from all $700$ training scenes that contain annotations for the class, and remove the annotations for all other classes.
If fewer than $300$ sequences are available for a class, we use only those.
Moreover, we determine the class ordering for the incremental steps based on the label counts, resulting in the order \textit{car} - \textit{pedestrian} - \textit{truck} - \textit{trailer} - \textit{motorcycle} - \textit{bicycle} - \textit{bus}.
For the group-incremental training procedure, we divide the data into disjoint sets of $234$, $233$, and $233$ sequences, which presents a challenging setting given the limited data.
Following COOLer~\cite{cit:ci_tracking-cooler}, we group similar classes into steps: We combine car, truck, bus, and trailer into the base step \textit{vehicle}, the class pedestrian into \textit{pedestrian} in the second step, and bicycle and motorcycle into \textit{two-wheeler} in the final step.

{\parskip=2pt
\noindent\textbf{Argoverse 2}}: %
We also create a class-incremental motion forecasting data split for the Argoverse 2 (AV2)~\cite{wilson2021argoverse} dataset, which contains traffic scenes from six US cities.
For all models, we resize the seven surround view RGB images to $640 \times 960$ pixels.
Since the front camera images are in portrait mode, we first crop them to landscape format.
Following Far3D~\cite{jiang2024far3d}, we split the RGB images of each of the $700$ sequences, recorded with 10 Hz, into five sequences with $2$ Hz.
We use the same 26 classes as in the official end-to-end forecasting evaluation.
For this dataset, we create a similar overlapping setting to that proposed by~\citet{cermelli2020modeling}, in which we use all available annotations for the classes at each step. We remove the annotations for classes from earlier and later steps, and, as a result, they may appear as background. For the base training, we randomly select 20 classes (\textit{articulated\_bus}, \textit{pedestrian}, \textit{construction\_cone}, \textit{bus}, \textit{wheelchair}, \textit{construction\_barrel}, \textit{motorcycle}, \textit{bicyclist}, \textit{dog}, \textit{stroller}, \textit{box\_truck}, \textit{sign}, \textit{stop\_sign}, \textit{wheeled\_device}, \textit{vehicular\_trailer}, \textit{mobile\_pedestrian\_crossing\_sign}, \textit{bicycle}, \textit{school\_bus}, \textit{truck\_cab}, \textit{regular\_vehicle}) and train the model for three incremental steps (20-2 setting), each with two of the remaining classes (\textit{motorcyclist}, \textit{wheeled\_rider} - \textit{truck}, \textit{bollard} - \textit{large\_vehicle}, \textit{message\_board\_trailer}).

{\parskip=2pt
\noindent\textbf{NeuroNCAP}}:
NeuroNCAP~\cite{ljungbergh2024neuroncap} is a neural rendering-based simulator to assess the performance of end-to-end planning methods in a closed-loop manner.
It contains 14 selected scenarios from the nuScenes validation set, modified to simulate stationary, frontal, and side collisions, for which it measures the collision rate and the NeuroNCAP score, which can reach a maximum of 5 stars.

\subsection{Baselines}
We compare \net~against several challenging baselines.
As it is desirable to come as close as possible to the performance of the \textit{joint training} that uses all sequences and all class labels simultaneously, this serves as the theoretical upper bound.
To demonstrate the difficulty of the proposed task, we benchmark a \textit{forgetting baseline} that does not employ any measures against forgetting in all proposed scenarios.
Further, to highlight the additional benefits of our approach beyond the pseudo-labeling from the object detections at future frames, we also report results for \textit{two pseudo-labeling baselines}.
The first uses forecasting predictions, and the second uses future 3D object detections via tracking IDs as pseudo-labels for forecasting.
In line with previous work, we adjust \textit{Learning without Forgetting (LwF)}~\cite{li2017learning} to our task and use it as a comparison. To be more specific, we employ knowledge distillation for classification and regression of the detection and motion prediction head. Since we do not have one-to-one correspondences between queries of the old and the new model due to temporal query propagation in the underlying end-to-end model, we can only perform knowledge distillation for positive pairs obtained via Hungarian matching with the pseudo-labels above. However, we do not add pseudo-labels to the ground truth labels. Instead, we employ the LwF soft classification loss and exclude them from the regular losses. This way, the model is ensured not to receive contradictory training signals for old classes.
Moreover, we adapt \textit{CL-DETR}~\cite{cit:ci_detection-cldetr}, a state-of-the-art method that combines pseudo-labeling and experience replay for class-incremental 2D object detection, for the presented task.
Similarly to our method, we limit the replay buffer size and retain only the samples per step that best represent the data distribution.
For the per-class incremental scenario, we adapt the sampling strategy to match the distribution of object frequencies across sequences.
To achieve a fair comparison, we use the pseudo-labels from future object detections for CL-DETR.
As an additional baseline for the exemplar replay buffer, we take inspiration from \textit{CoDEPS}~\cite{voedisch23codeps}, which stores exemplar samples based on their feature similarity for online continual depth estimation and panoptic segmentation.
We adopt this idea by using the global classification feature vector from DINOv3~\cite{simeoni2025dinov3} for each input sample, stacking these vectors for all samples in a sequence, and iteratively selecting sequences for the buffer that have the lowest pair-wise Chamfer cosine similarity to all sequences in the buffer.
To follow CoDEPS, we compute cosine similarity rather than a distance function.

\begin{table}
\scriptsize
\centering
\caption{Class-incremental open-loop planning results after each training stage on the nuScenes validation set for the proposed per-class incremental setting. The best results are marked \textbf{bold}, and the second-best \underline{underlined}.\label{tab:planning}}
\begin{tabular}{c|ccc>{\columncolor[gray]{0.9}}c|ccc>{\columncolor[gray]{0.9}}c} 
    \toprule
    \multirow{2.5}{*}{Step} & \multicolumn{4}{c|}{L2 $(m)\downarrow$} 
    & \multicolumn{4}{c}{Col. Rate $(\%)\downarrow$}\\
    \cmidrule{2-9}
    & 1$s$ & 2$s$ & 3$s$ & Avg.
    & 1$s$ & 2$s$ & 3$s$ & Avg.\\
    \midrule
    Joint Training & 0.32 & 0.63 & 1.04 & 0.66 & 0.02 & 0.08 & 0.34 & 0.14\\
    \midrule\midrule
    1 & 0.38 & 0.76 & 1.27 & 0.80 & 0.03 & 0.12 & 0.60 & 0.25\\
    2 & 0.36 & 0.71 & 1.18 & 0.75 & \underline{0.02} & 0.10 & 0.45 & 0.19\\
    3 & \underline{0.33} & 0.64 & 1.05 & \underline{0.67} & \underline{0.02} & 0.10 & 0.41 & 0.18\\
    4 & 0.35 & 0.67 & 1.10 & 0.70 & 0.03 & 0.16 & 0.46 & 0.21\\
    5 & 0.34 & 0.65 & 1.06 & 0.68 & \underline{0.02} & \underline{0.09} & \textbf{0.31} & \textbf{0.14}\\
    6 & \underline{0.33} & \underline{0.63} & \underline{1.04} & \underline{0.67} & \textbf{0.01} & \textbf{0.08} & \underline{0.34} & \underline{0.15}\\\midrule
    7 & \textbf{0.32} & \textbf{0.63} & \textbf{1.03} & \textbf{0.66} & \underline{0.02} & 0.11 & 0.36 & 0.16\\
    \bottomrule
\end{tabular}
\end{table}

\begin{figure}
\centering
\includegraphics[width=\linewidth]{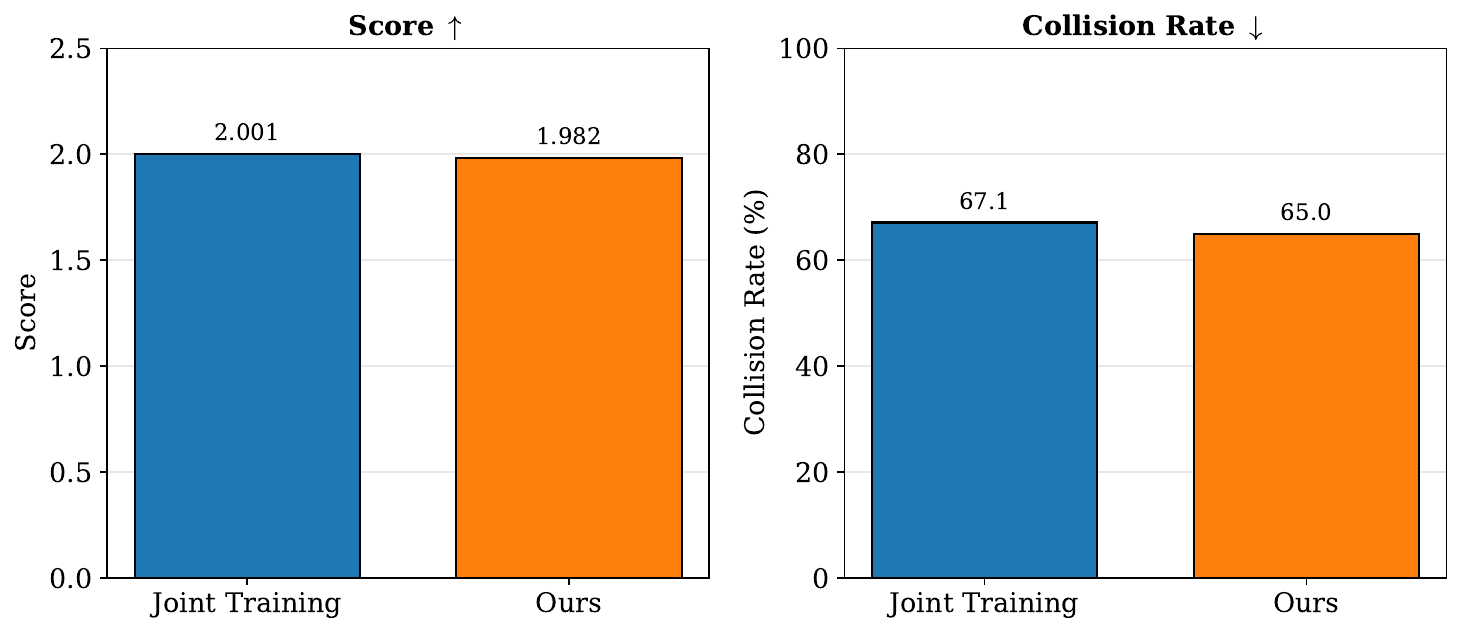}
\caption{Class-incremental closed-loop planning results on the NeuroNCAP evaluation for the proposed per-class incremental setting. Compared to jointly training the model with all classes, \net~achieves a comparable NeuroNCAP score and an even slightly lower collision rate.}\label{fig:planning_neuroncap}
\end{figure}

\begin{table*}
\scriptsize
\centering
\caption{Ablation study on the per-class incremental setting to study the influence of each contribution. The best results are marked \textbf{bold}.\label{tab:ablation_contributions}}
\begin{tabular}{cccc|ccccccc|>{\columncolor[gray]{0.9}}c} 
    \toprule
    \multicolumn{2}{c}{Forecasting PL} & \multirow{3}{*}{\shortstack[l]{VLM\\Filt.}} & \multirow{3}{*}{\shortstack[l]{Var.\\Buffer}} & \multicolumn{7}{c}{AMOTA (\%)$\uparrow$ / mAP\_f (\%)$\uparrow$}\\
    \cmidrule{5-12}
    pred. & fut det. & & & (1) & (2) & (3) & (4) & (5) & (6) & (7) & All\\
    \midrule
    \multicolumn{2}{l}{Joint Training} & & & 49.9 / 32.75 & 32.4 / 25.73 & 29.5 / 15.15 & 0.0 / 5.14 & 32.6 / 17.81 & 36.8 / 24.19 & 33.8 / 18.34 & 30.7 / 19.87\\
    \midrule\midrule
    & & & & 0.0 / 0.0 & 0.0 / 0.0 & 0.0 / 0.0 & \textbf{0.0} / 0.0 & 0.0 / 0.0 & 0.0 / 0.0 & \textbf{36.4} / 12.39 & 5.2 / 1.77\\
    \cmark & & & & 29.2 / 18.19 & 17.4 / 13.76 & 5.2 / 6.71 & \textbf{0.0} / 1.50 & 25.2 / 11.01 & 25.0 / 15.08 & 33.8 / \textbf{16.36} & 19.4 / 11.80\\
    & \cmark & & & 41.3 / 27.27 & 20.8 / 14.50 & 11.0 / 9.04 & \textbf{0.0} / 3.48 & 18.7 / 9.75 & 27.2 / 17.82 & 26.2 / 14.21 & 20.8 / 13.73\\
    & \cmark & \cmark & & 52.0 / 29.45 & 30.4 / 14.70 & \textbf{19.6} / 10.34 & \textbf{0.0} / 3.10 & \textbf{27.8} / 12.65 & 32.1 / 17.90 & 30.2 / 14.06 & 27.4 / 14.60\\
    & \cmark & \cmark & \cmark & \textbf{52.7} / \textbf{29.95} & \textbf{32.5} / \textbf{15.82} & 15.8 / \textbf{11.26} & \textbf{0.0} / \textbf{3.85} & 26.3 / \textbf{13.46} & \textbf{33.9} / \textbf{19.62} & 35.2 / 15.27 & \textbf{28.0} / \textbf{15.60}\\
    \bottomrule
\end{tabular}
\end{table*}

\begin{figure*}[!ht]
    \centering

    \begin{subfigure}{\linewidth}
        \centering
        \includegraphics[width=0.81\linewidth]{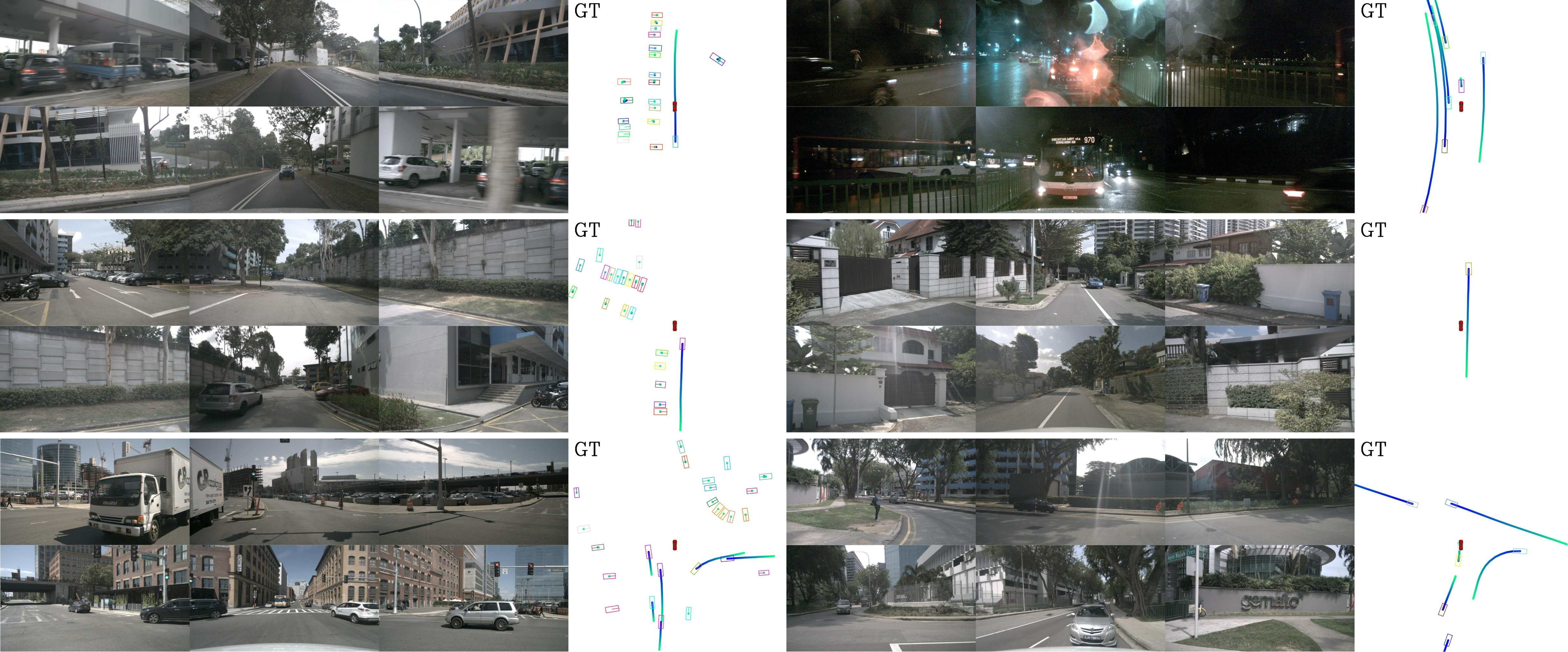}
        \caption{CL-DETR~\cite{cit:ci_detection-cldetr}.}
    \end{subfigure}

    \vspace{0.5em} %

    \begin{subfigure}{\linewidth}
        \centering
        \includegraphics[width=0.81\linewidth]{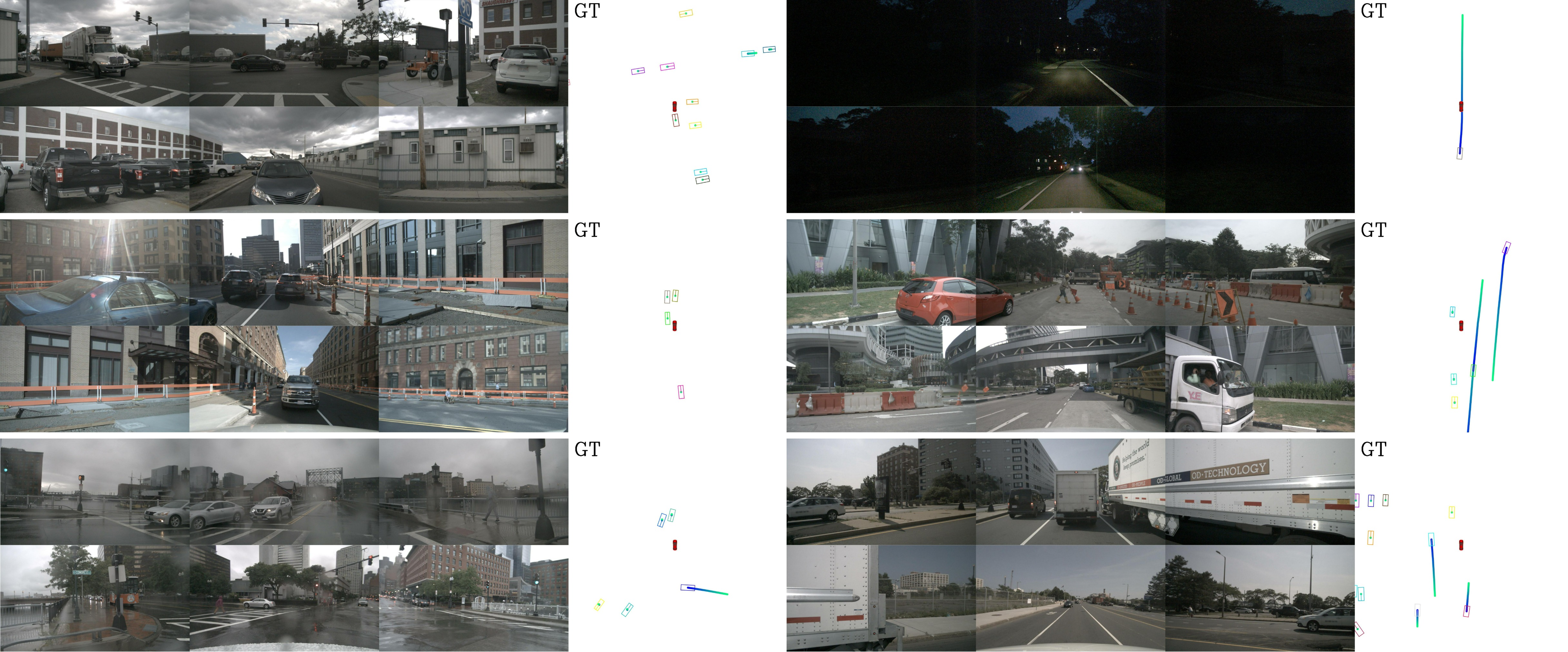}
        \caption{DINOv3.}
    \end{subfigure}

    \vspace{0.5em} %

    \begin{subfigure}{\linewidth}
        \centering
        \includegraphics[width=0.81\linewidth]{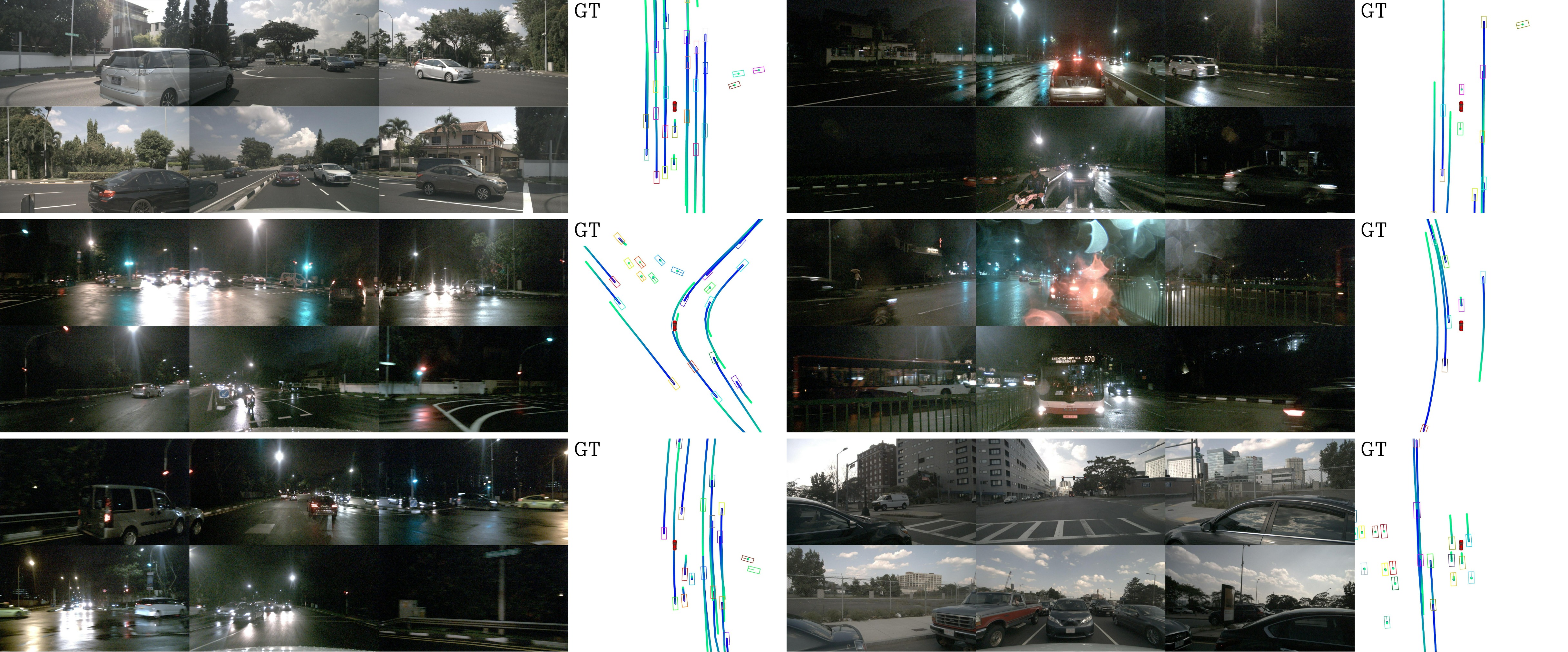}
        \caption{Ours.}
    \end{subfigure}

    \caption{Qualitative comparison of selected scenes per replay buffer strategy (CL-DETR~\cite{cit:ci_detection-cldetr}, DINOv3, Ours). We show one sample from each of the six highest-rated sequences per method after training the class \textit{car} on the per-class incremental setting. As can be seen, the sequences selected by CL-DETR represent the data label distribution of the dataset well, but only contain a few moving objects without particularly interesting motion patterns. When using diverse DINO features, the scenes cover a variety of weather and lighting conditions in the image space, but they often only contain a few objects that are not necessarily moving. In contrast, our proposed sampling strategy selects scenarios with a large number of moving objects with diverse motions.}
    \label{fig:replay}
\end{figure*}

\begin{table}
\scriptsize
\centering
\caption{Ablation study for the exemplar selection strategy used on both proposed nuScenes settings. For the random exemplar selections, we run three trials with different random seeds and average the results. The best results are marked \textbf{bold}.\label{tab:ablation_buffer_selection}}
\begin{tabular}{l|l|cc>{\columncolor[gray]{0.9}}c|cc>{\columncolor[gray]{0.9}}c} 
    \toprule
    Setting & Selection & \multicolumn{3}{c|}{AP\_f$_{linear}$ (\%)$\uparrow$} 
    & \multicolumn{3}{c}{AP\_f$_{non-linear}$ (\%)$\uparrow$}\\
    \midrule
    \multirow{6}{*}{\shortstack[l]{per-class\\(7 steps)}} & Classes & (1) & (2) & All
    & (1) & (2)& All\\
    \cmidrule{2-8}
    & Random & 27.32 & 19.26 & 15.68 & 12.90 & 6.37 & 4.29\\
    & DINOv3 & 26.97 & 18.99 & 15.25 & 12.75 & 6.69 & 4.01\\
    & CL-DETR & 27.04 & 19.28 & 14.58 & 12.86 & 6.80 & 4.10\\
    & Ours & \textbf{28.86} & \textbf{19.50} & \textbf{16.24} & \textbf{13.04} & \textbf{8.10} & \textbf{4.57}\\
    \midrule\midrule
    \multirow{6}{*}{\shortstack[l]{group\\(3 steps)}} & Classes & (1-4) & (5) & All & (1-4) & (5) & All\\
    \cmidrule{2-8}
    & Random & 7.32 & 18.75 & 9.91 & 2.67 & 6.79 & \textbf{2.58}\\
    & DINOv3 & \textbf{7.71} & 18.94 & 10.29 & 2.20 & \textbf{6.89} & 2.31\\
    & CL-DETR & 6.08 & 19.21 & 9.36 & 2.11 & 7.82 & 2.43\\
    & Ours & 7.18 & \textbf{20.75} & \textbf{10.69} & \textbf{2.75} & 6.44 & 2.54\\
    \bottomrule
\end{tabular}
\end{table}

\begin{table}[ht]
\scriptsize
\centering
\caption{Ablation study for the size of the replay buffer $K$ on the per-class incremental setting. Note that we keep the buffer size constant across steps. The best results are marked \textbf{bold}.\label{tab:ablation_replay_size}}
\begin{tabular}{l|ccc|ccc} 
    \toprule
    \multirow{2.5}{*}{Buffer Size} & \multicolumn{3}{c|}{AP\_f$_{non-linear}$ (\%)$\uparrow$}
    & \multicolumn{3}{c}{mAP\_f (\%)$\uparrow$}\\
    \cmidrule{2-7}
    & (1) & (2) & All
    & (1) & (2) & All\\
    \midrule
    $K=6$ & 11.84 & 5.53 & 3.56 & 29.60 & 15.10 & 14.86\\
    $K=12$ & 12.53 & 7.55 & 4.37 & \textbf{30.16} & 15.14 & 15.50\\
    $K=30$ & \textbf{13.04} & \textbf{8.10} & \textbf{4.57} & 29.95 & \textbf{15.82} & \textbf{15.60}\\
    \bottomrule
\end{tabular}
\end{table}

\begin{table}[ht]
\scriptsize
\centering
\caption{Ablation study for the statistical function used to select the replay buffer samples based on the latent query space on the per-class incremental setting. The best results are marked \textbf{bold}.\label{tab:abl_statistical_function}}
\begin{tabular}{l|ccc|ccc} 
    \toprule
    \multirow{2.5}{*}{Function} & \multicolumn{3}{c|}{AP\_f$_{non-linear}$ (\%)$\uparrow$}
    & \multicolumn{3}{c}{mAP\_f (\%)$\uparrow$}\\
    \cmidrule{2-7}
    & (1) & (2) & All
    & (1) & (2) & All\\
    \midrule
    STD & 12.62 & 6.61 & 4.48 & 29.54 & 14.55 & 15.13\\
    MAD & \textbf{13.23} & 6.77 & 4.34 & \textbf{29.97} & 14.52 & 15.00\\
    SS & 13.04 & \textbf{8.10} & \textbf{4.57} & 29.95 & \textbf{15.82} & \textbf{15.60}\\
    \bottomrule
\end{tabular}
\end{table}

\begin{table}[ht]
\scriptsize
\centering
\caption{Ablation study for the pseudo-labeling threshold $\theta$ on the per-class incremental setting. The best results are marked \textbf{bold}.\label{tab:abl_pseudo_labeling_thresh}}
\begin{tabular}{l|ccc|ccc} 
    \toprule
    \multirow{2.5}{*}{Threshold} & \multicolumn{3}{c|}{AP\_f$_{non-linear}$ (\%)$\uparrow$}
    & \multicolumn{3}{c}{mAP\_f (\%)$\uparrow$}\\
    \cmidrule{2-7}
    & (1) & (2) & All
    & (1) & (2) & All\\
    \midrule
    $\theta=0.2$ & 13.02 & 7.13 & 3.97 & 28.90 & 14.86 & 14.58\\
    $\theta=0.3$ & \textbf{13.04} & \textbf{8.10} & \textbf{4.57} & \textbf{29.95} & \textbf{15.82} & \textbf{15.60}\\
    $\theta=0.4$ & 12.20 & 7.11 & 4.42 & 28.46 & 13.63 & 15.00\\
    \bottomrule
\end{tabular}
\end{table}

\begin{table*}[ht]
\scriptsize
\centering
\caption{Comparison of the training time $t$ between \net~and the joint re-training with all labels in each step (0-2) for the group-incremental nuScenes setting, divided into training time $t_\text{train}$, replay inference \& buffer generation time $t_\text{replay}$, and pseudo labeling time $t_\text{pseudo}$.\label{tab:time_comp}} %
\begin{tabular}{l|ccc|ccc|c|>{\columncolor[gray]{0.9}}c} 
    \toprule
    \multirow{2.5}{*}{Method}
    & \multicolumn{3}{c|}{Step 0 $(hh:mm)\downarrow$} 
    & \multicolumn{3}{c|}{Step 1 $(hh:mm)\downarrow$} 
    & Step 2 $(hh:mm)\downarrow$
    & Total $(hh:mm)\downarrow$\\
    \cmidrule{2-9}
    & $t_\text{train}$ & $t_\text{replay}$ & $t_\text{pseudo}$ & $t_\text{train}$ & $t_\text{replay}$ & $t_\text{pseudo}$ & $t_\text{train}$ & $t_\text{total}$\\
    \midrule
    Joint Training & 03:17 & - & - & 06:41 & - & - & 09:59 & 19:57\\
    Ours w/o pre-computed VLM masks & 03:17 & 00:04 & 01:18 & 03:21 & 00:04 & 01:18 & 03:08 & \underline{12:30}\\
    Ours w/ pre-computed VLM masks & 03:17 & 00:04 & 00:09 & 03:21 & 00:04 & 00:09 & 03:08 & \textbf{10:12}\\
    \bottomrule
\end{tabular}
\vspace{-0.1cm}
\end{table*}

\begin{table}[ht]
\scriptsize
\centering
\caption{Ablation study between Grounded SAM 2 and a light-weight open-vocabulary segmentation model (YOLOE-26l-seg) on the per-class incremental setting. The best results are marked \textbf{bold}.\label{tab:abl_VLM}}
\begin{tabular}{l|ccc|ccc} 
    \toprule
    \multirow{2.5}{*}{VLM} & \multicolumn{3}{c|}{AMOTA (\%)$\uparrow$}
    & \multicolumn{3}{c}{mAP\_f (\%)$\uparrow$}\\
    \cmidrule{2-7}
    & (1) & (2) & All
    & (1) & (2) & All\\
    \midrule
    YOLOE-26l-seg~\cite{wang2025yoloe} & 52.1 & 31.3 & 27.8 & \textbf{30.18} & 15.53 & 15.59\\
    Grounded SAM 2~\cite{ren2024grounded} & \textbf{52.7} & \textbf{32.5} & \textbf{28.0} & 29.95 & \textbf{15.82} & \textbf{15.60}\\
    \bottomrule
\end{tabular}
\end{table}

\begin{table}[ht]
\scriptsize
\centering
\caption{Ablation study for the online map generation (OMG) module on the per-class incremental setting. The best results are marked \textbf{bold}.\label{tab:online_map_generation}}
\begin{tabular}{c|ccc|ccc} 
    \toprule
    \multirow{2.5}{*}{OMG} & \multicolumn{3}{c|}{AP\_f$_{non-linear}$ (\%)$\uparrow$}
    & \multicolumn{3}{c}{mAP\_f (\%)$\uparrow$}\\
    \cmidrule{2-7}
    & (1) & (2) & All
    & (1) & (2) & All\\
    \midrule
    \cmark & 11.94 & 6.68 & 3.52 & 28.27 & 15.09 & 14.31\\
    \xmark & \textbf{13.04} & \textbf{8.10} & \textbf{4.57} & \textbf{29.95} & \textbf{15.82} & \textbf{15.60}\\
    \bottomrule
\end{tabular}
\vspace{-0.2cm}
\end{table}

\begin{table}[ht]
\scriptsize
\centering
\caption{Comparison between a class-agnostic motion forecasting algorithm and \net~on the per-class incremental setting. For this experiment, the motion forecasting head is frozen after training on cars. The best results are marked \textbf{bold}. The results are shown for the class truck (3) that has similar motion behavior as cars, and the classes pedestrian (2) and bicycle (6) with more dissimilar behavior.\label{tab:abl_class_agnostic}}
\begin{tabular}{c|cccc|cccc} 
    \toprule
    \multirow{2.5}{*}{\shortstack[l]{Class-\\aware}} & \multicolumn{4}{c|}{AP\_f$_{non-linear}$ (\%)$\uparrow$}
    & \multicolumn{4}{c}{mAP\_f (\%)$\uparrow$}\\
    \cmidrule{2-9}
    & (3) & (2) & (6) & All
    & (3) & (2) & (6) & All\\
    \midrule
    \xmark & 2.29 & 3.12 & 0.06 & 3.18 & 10.08 & 10.83 & 17.25 & 14.51\\
    \cmark & \textbf{2.91} & \textbf{8.10} & \textbf{0.14} & \textbf{4.57} & \textbf{11.26} & \textbf{15.82} & \textbf{19.62} & \textbf{15.60}\\
    \bottomrule
\end{tabular}
\vspace{-0.2cm}
\end{table}

\subsection{Training Protocol}
To develop our approach, we build on the end-to-end prediction and planning method SparseDrive~\cite{cit:prediction_e2e-sparsedrive}.
Unless otherwise noted, we use the same parameters as in the SparseDrive-S model and train it with the same loss functions.
In contrast to the original work, we use only camera images as input for training and testing, and we disable LiDAR-based depth supervision throughout training to develop a camera-only method.
In addition, we remove the online map generation module, as its influence on end-to-end motion forecasting models remains underexplored~\cite{xu2024towards}, as also shown in~\Cref{subsec:ablations}, and deactivate ego-trajectory planning for most of the experiments.
In each incremental step, we train the model in two stages as in~\cite{cit:prediction_e2e-sparsedrive}, using the AdamW optimizer~\cite{loshchilov2018decoupled} with cosine learning rate decay and weight decay of $1 \times 10^{-3}$.
In the first stage, we train only the 3D object detection model for $50$ epochs on nuScenes and $10$ on AV2. In the second stage, we update the entire model for $10$ and $2$ epochs, respectively.
To enhance plasticity for novel classes, we only use replay samples during the second-stage training, with each batch containing approximately 25\% replay samples.
We set the initial learning rates for the first and second stages to $2 \times 10^{-4}$ and $1 \times 10^{-4}$ during base training, and to half of these values in the subsequent incremental steps.
For the first stage, we enable query denoising from Sparse4Dv3~\cite{lin2023sparse4d}, but only for the ground-truth annotations due to possible noise in the pseudo-labels.
For nuScenes, we train all models on $4$ GPUs with a total batch size of $64$ for the first stage and $48$ for the second stage, while setting these numbers to $32$ and $28$ for AV2.

To train the model on the AV2 dataset, we follow the velocity calculation and trajectory GT generation from LT3D~\cite{xu2024towards} for the AV2 End-to-End Forecasting Challenge and extend the dataloader from Far3D~\cite{jiang2024far3d} for SparseDrive.
Following Far3D, we use VoVNet-99~\cite{lee2019energy} as the image backbone for AV2, while employing ResNet-50~\cite{he2016deep} for all models on nuScenes.
We forecast the object motion for a time horizon $T$ of $6$ s on the nuScenes dataset, as in prior work~\cite{hu2023planning, cit:prediction_e2e-vip3d, cit:prediction_e2e-sparsedrive}, and for $3$ s on the AV2 dataset, as in the End-to-End Forecasting Challenge.
Regarding the specific parameters for the incremental learning scenarios, we set the fixed replay buffer size $N_{replay}$ to $30$ sequences across all experiments, unless otherwise specified.
If the buffer size is not divisible without a residual by the number of classes, we select the remaining sequences from the latest sub-dataset.
If there is a residual again, we store more samples from the classes with the lowest frequencies.
To create the pseudo-labels, we use a threshold of $\theta = 0.3$, unless stated explicitly.

\subsection{Quantitative Results}
{\parskip=2pt
\noindent\textbf{Motion Forecasting:}
For each method, we compute the forecasting mean average precision (mAP\_f)~\cite{Peri_2022_CVPR} as the main metric, which was proposed to evaluate end-to-end motion forecasting approaches.
It averages static, linear, and non-linear Forecasting Average Precision (AP\_f), which we also report separately.
For nuScenes, we adapt the evaluation code from the AV2-API.

\subsubsection{nuScenes Per-Class Incremental Setting}
First, we present the forecasting results for the nuScenes per-class incremental split in \Cref{tab:quant_results_per_class}.
In this setting, \net~achieves the highest overall mAP\_f and is especially more accurate for the classes introduced first, which are more prone to forgetting.
Moreover, the accuracy for linearly and non-linearly moving objects, which are more complex for the model to learn, significantly surpasses that of the baselines.
It has to be emphasized that due to the training with only 300 sequences per step, it is not expected that the incremental methods are able to achieve the performance of the upper bound of joint training.
Since the focus of our method is to retain knowledge about these moving objects, there is a slight trade-off regarding static objects, while CL-DETR, which is based on dataset distributions, exhibits degraded performance for moving agents.
We also make a similar observation for the 3D object detection and tracking metrics in \Cref{tab:quant_results_det}: While CL-DETR~\cite{cit:ci_detection-cldetr} achieves a slightly higher mean average precision (mAP) and nuScenes detection score (NDS)~\cite{caesar2020nuscenes} for the first introduced class car, \net~still tracks object of this class better, as can be seen by the average multi-object tracking accuracy (AMOTA)~\cite{caesar2020nuscenes}.
Moreover, it substantially outperforms the baselines on detection and tracking metrics when averaged across all classes, which we attribute to the VLM-guided filtering mechanism, as shown in the ablation studies in \Cref{subsec:ablations}.
For this data split, we also evaluate the end-to-end prediction accuracy (EPA)~\cite{cit:prediction_e2e-vip3d} in \Cref{tab:quant_results_per_class}, which shows very low scores for methods that do not counteract increased model confidence over time.
The reason for this is that, contrary to mAP\_f, EPA does not consider the relative score ordering of objects but strictly punishes false positives above a certain confidence threshold.
To make the evaluation more unbiased, we select mAP\_f as the main metric for the remaining experiments.

\subsubsection{nuScenes Group-Incremental Setting}
In the group-incremental scenario, we observe that this setting is very challenging due to the smaller dataset size, which leads to lower accuracies across all methods, as demonstrated in \Cref{tab:quant_results}.
Nevertheless, \net~achieves the best results for static and linear trajectories, while the accuracy for non-linearly moving objects is on a low level for all methods.

\subsubsection{AV2}
On the proposed overlapping AV2 scenario, \net~also outperforms the baselines in motion prediction, primarily for moving objects, as shown in \Cref{tab:quant_results_av2}.
Notably, it nearly matches the results of joint training, even surpassing them for linearly moving objects.
The narrower gap in this setting is due to the shorter forecasting horizon and the larger number of training sequences compared to the experiments on nuScenes.

\subsubsection{Forgetting}
In addition to evaluating the absolute performance after the incremental training, we compute forgetting percentage points (FPPs)~\cite{cit:ci_detection-cldetr} for mAP\_f to distinctively evaluate forgetting in~\Cref{tab:forgetting_metrics}.
This metric evaluates the difference in mAP\_f for the first set of classes $C^0$ between the base training and the final incremental step.
While most methods experience a drop in performance, \net~improves its forecasting performance on the nuScenes dataset and shows only slight forgetting on AV2, achieving the least forgetting among all methods.
The reason our method can achieve negative FPPs is the coupling between agents for forecasting: After the first stage, the model has not learned to perceive all agent classes, which can influence the future movements of agents in class $C^0$.

In~\Cref{fig:results_per_class}, we visualize the motion prediction metrics per class for \net~over the course of incremental steps for the per-class incremental setting.
We observe that the classes introduced in later steps are more prone to forgetting, likely due to reduced plasticity in the model and a lower number of samples for these classes in the dataset, given our selection order for the incremental steps.

{\parskip=2pt
\noindent\textbf{Planning:}
\setcounter{subsubsection}{0}\subsubsection{nuScenes}
In \Cref{tab:planning}, we show that our approach can be successfully applied to open-loop planning scenarios on the nuScenes dataset, in which the model learns novel agent types over time.
As a result, across the added classes, it can be observed that the L2 Error for the ego future trajectory and the collision rate gradually decrease.

\subsubsection{NeuroNCAP}
In \Cref{fig:planning_neuroncap}, we assess the model's capability for closed-loop driving on the NeuroNCAP~\cite{ljungbergh2024neuroncap} benchmark in comparison with the joint training with all classes.
Following prior work~\cite{ljungbergh2024neuroncap, zhang2025bridging}, we perform a post-processing step similar to UniAD~\cite{hu2023planning}, in which we use the predicted object detections and future trajectories for collision avoidance.
We observe that both models achieve similar scores and collision rates, demonstrating that our incremental training can match the planning performance of joint training.

\begin{figure*}[!ht]
    \centering
    {\includegraphics[width=0.9\linewidth]{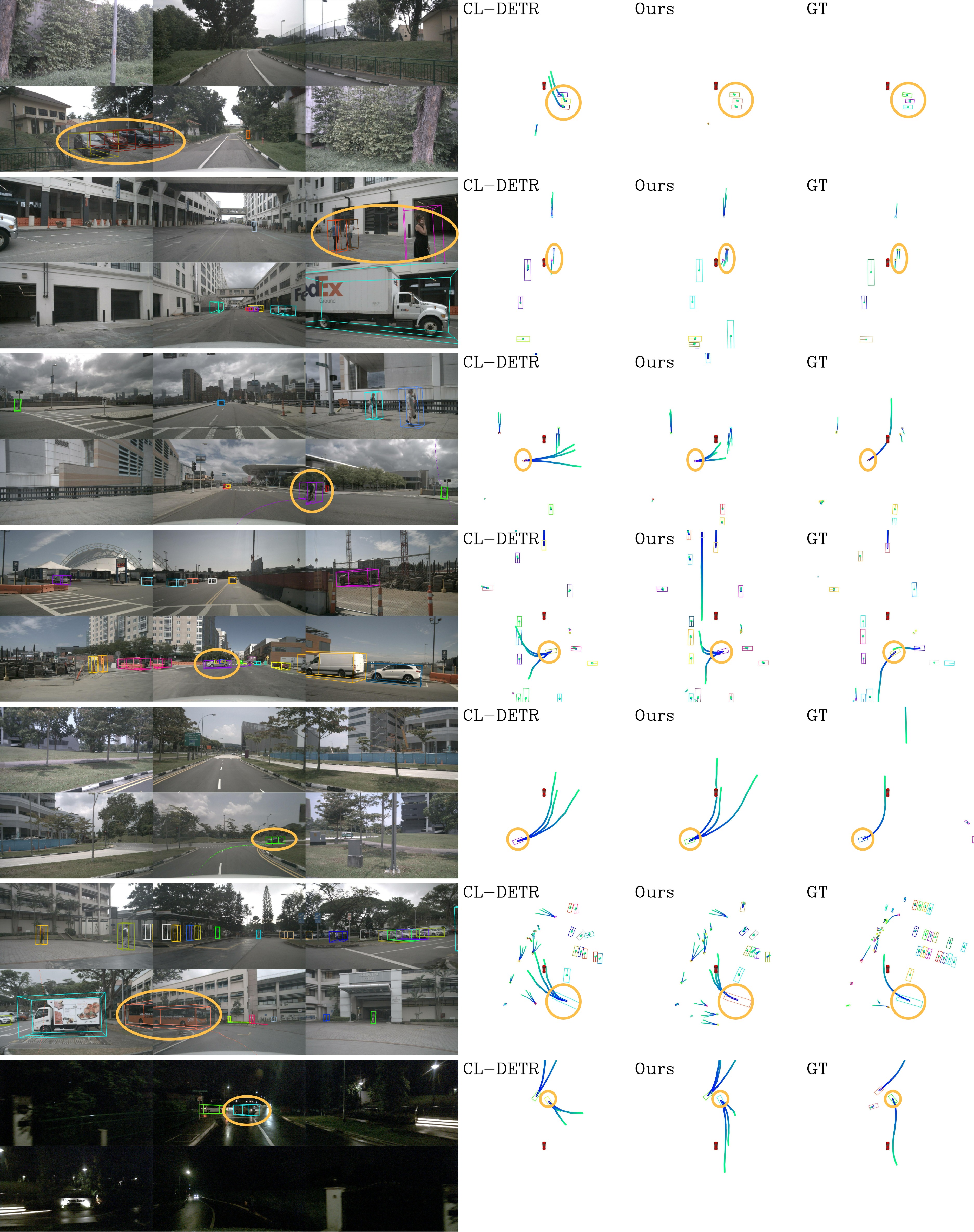}}
    \caption{Qualitative results on the nuScenes dataset for the proposed per-class incremental setting. The range of the birds-eye-view images is set to \SI{40}{\meter} in the longitudinal and lateral direction of the ego-vehicle. In the top two examples, it can be seen that CL-DETR~\cite{cit:ci_detection-cldetr} might predict movements for static objects and more often misses objects such as pedestrians than \net. Moreover, as seen in the residual rows, our approach predicts non-linear motions for objects of different classes more accurately, also under varying lighting conditions.}
    \label{fig:nuscenes_ext}
\end{figure*}

\begin{figure*}[!ht]
    \centering
    {\includegraphics[width=\linewidth]{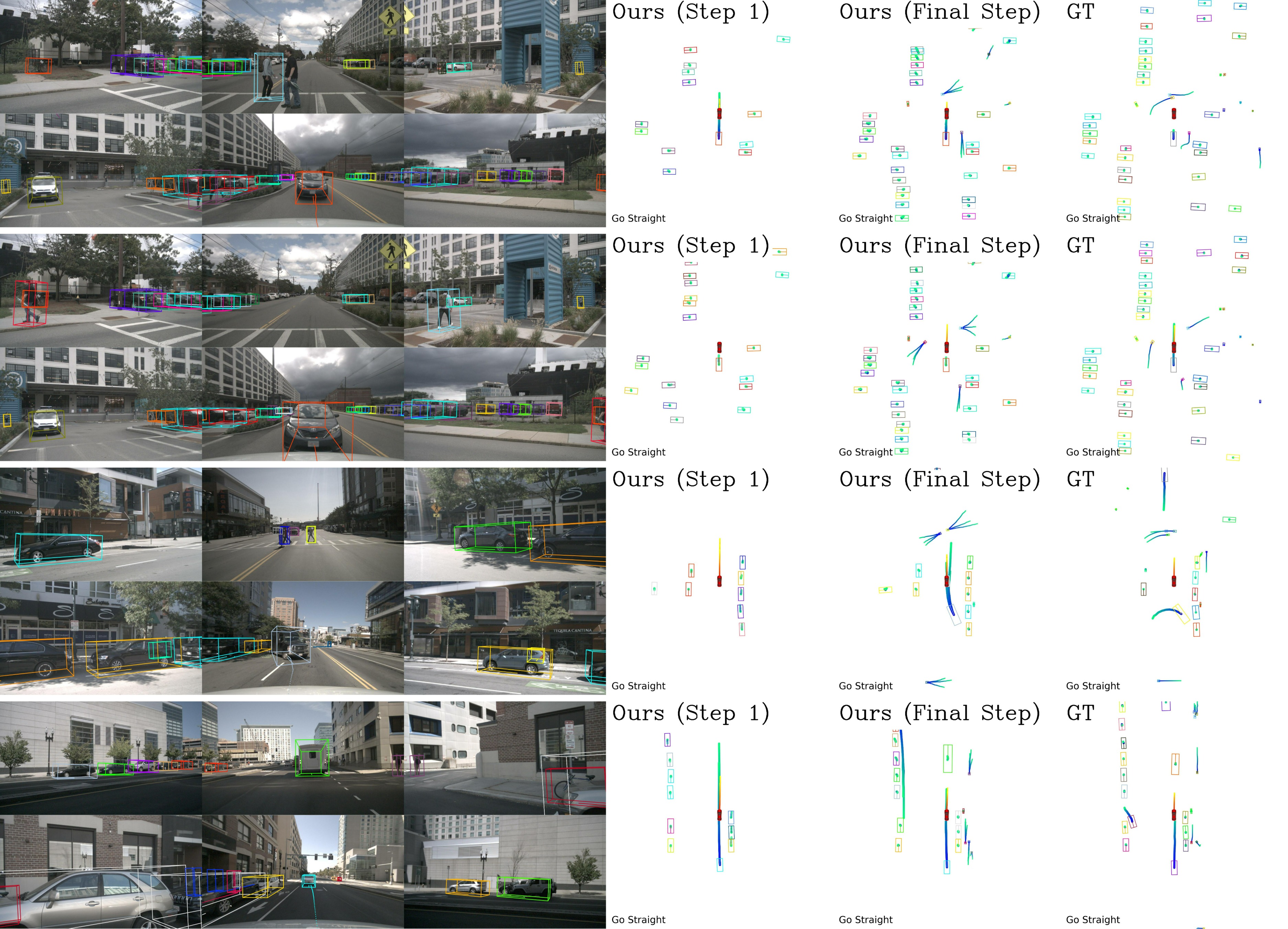}}
    \caption{Qualitative open-loop planning results for the per-class incremental setting on the nuScenes dataset. The top-$3$ modes for the planned future ego-trajectory over the next \SI{6}{\second} are shown in orange from a birds-eye view, while the planning command is printed at the bottom left. The range of the birds-eye-view images is set to \SI{40}{\meter} in the longitudinal and lateral direction of the ego-vehicle. We compare the results from the first step of training (only with cars) against the final trained model and the ground truth (GT).}
    \label{fig:planning}
\end{figure*}

\subsection{Ablation Study}\label{subsec:ablations}
We investigate the effect of each individual contribution of the approach in \Cref{tab:ablation_contributions}.
First, creating motion forecasting pseudo-labels from object detections in future frames greatly increases mAP\_f.
Second, the VLM-based filtering significantly improves tracking and also positively impacts forecasting.
Third, the sequence-based experience replay further boosts motion forecasting results.

We also compare the variance-based replay mechanism directly with other replay buffer population strategies in \Cref{tab:ablation_buffer_selection} for the two nuScenes splits.
Here, we conclude that our approach is especially effective in retaining knowledge about moving agents compared to methods that consider only the image feature space or dataset distributions.
We visually compare the sequences that different replay buffer strategies select after the base training for the first class \textit{car} for the nuScenes per-class incremental setting in \Cref{fig:replay}.
For visualization purposes, we show only the six sequences with the highest scores, i.e., the sequences selected first by the corresponding algorithms.
Since the DINOv3~\cite{simeoni2025dinov3}-based strategy, inspired by~\cite{voedisch23codeps}, favors diverse sequences in the image feature space, the respective buffer contains sequences with atypical scenarios, such as construction zones, varying lighting and weather conditions, and from different cities.
Although this covers the image feature space well, it yields worse motion forecasting results than our method, as shown in \Cref{tab:ablation_buffer_selection}.
Moreover, the sequences selected by CL-DETR~\cite{cit:ci_detection-cldetr} aim to optimally represent the data-label distribution.
Therefore, the selected sequences generally have approximately the same distribution across object classes as the whole dataset.
The first selected sequences show a few turning maneuvers, but overall, they contain many static objects, as is the case across the whole dataset.
In contrast to these methods, we demonstrate that our proposed strategy, which considers the motion query latent space, selects sequences with a particularly high number of moving objects.
As illustrated in the figure, the selected sequences exhibit dense traffic and multiple cars moving or turning.
The selected scenes also include particularly interesting scenarios for motion forecasting, such as mixtures of moving and stopped cars or accelerating cars after a traffic light turns green, as shown in the bottom-right image. 
Furthermore, we present ablations on the replay buffer size $K$ in \Cref{tab:ablation_replay_size}, which indicate that storing only two sequences per class in the final step ($K=12$) already notably enhances motion forecasting quality.

In \Cref{tab:abl_statistical_function}, we compare different statistical functions - standard deviation (STD), mean average deviation (MAD), and sum of squared deviations (SS) - to compute the score for the replay selection.
Using the sum of squared deviations yields the best results because it places greater weight on scenes with many dynamic objects.
The other methods are more prone to selecting scenes featuring a single dynamic object with atypical motion, which provides less training signal to mitigate forgetting. 
Furthermore, we ablate the pseudo-labeling threshold $\theta$ in \Cref{tab:abl_pseudo_labeling_thresh}. Using a value of $0.3$ provides the best balance between using too many and too few predictions as pseudo-labels.

If the offline process of our pseudo-labeling mechanism were to run on an edge device, it would be beneficial to use a lighter, open-vocabulary segmentation model rather than Grounded SAM 2. %
Therefore, we demonstrate in \Cref{tab:abl_VLM} that \net~also maintains its performance when the much smaller YOLOE-26l-seg model is used to provide instance masks for the VLM-guided pseudo-label filtering.

Since map priors are intended to enhance motion forecasting results and may reduce forgetting, we enable the online map generation (OMG) head of SparseDrive for incremental training in \Cref{tab:online_map_generation}.
Here, we observe that the model's forecasting performance degrades significantly. The reasons for this behavior are concurrent training signals and the underlying model's inadequate exploitation of the additional information.
The finding that map information is underutilized in recent end-to-end models such as SparseDrive is consistent with prior research~\cite{xu2024towards}.

Finally, we justify the class-incremental setup for end-to-end motion prediction by comparing \net~to a model for which the motion prediction head is only trained during base training and kept frozen afterward in \Cref{tab:abl_class_agnostic}.
Although this class-agnostic motion predictor can learn useful kinematic priors for similar classes, it fails when confronted with more distinct agent types.
In addition, it experiences more forgetting of the base class than our approach does, which can be attributed to the internal representation of the object detection head shifting over the course of incremental training to better discriminate between classes, while the motion prediction head must remain frozen.

\subsection{Training and Inference Time}
\label{sec:training_time}

To further highlight the benefits of our approach, we compare its training time in the group-incremental setting with that of joint training and refinement, which uses all available labels and sequences at each incremental step, in \Cref{tab:time_comp}.
We run this experiment with eight NVIDIA A6000, each with 48 GB of video random-access memory (VRAM), on a server with 256 GB of random-access memory (RAM) and an AMD EPYC 7452 32-core processor.

As illustrated in the table, the total time to iteratively retrain the model with all available labels is almost twice that of \net~when assuming that Grounded SAM 2~\cite{ren2024grounded} masks for the old classes are provided for each novel data subset.
This assumption might be valid, since the masks can be precomputed outside the edge device due to their independence from the individual model.
However, if this assumption is not made and the time for mask generation is included in the total required time, the incremental training is still about 40\% faster.
The other framework components, aside from VLM-based mask generation, incur almost no additional computational overhead: To generate the replay buffer, only a forward pass through the previous dataset $D^{i-1}$ is needed to retrieve the latent queries.
Similarly, pseudo-label generation requires only a forward pass through the new dataset $D^{i}$ but takes slightly longer due to the sequential matching of objects to VLM masks. 
It should also be noted that, to enable the joint training, it would be necessary to relabel all past sequences with the novel class and provide annotations for all past classes in the novel data subset, in addition to the time required for training.
Consequently, this process would require significant additional effort beyond the raw training time.

Since our adaptations focus on two offline processes, an improved pseudo-label quality and a replay selection strategy, they do not modify the architectural components of the end-to-end driving model itself.
Consequently, our final trained model achieves the same runtime during inference as the underlying SparseDrive~\cite{cit:prediction_e2e-sparsedrive}.

\subsection{Qualitative Results}
\subsubsection{Motion Forecasting}
We present qualitative results of the motion forecasting results from our method, compared with those from CL-DETR~\cite{cit:ci_detection-cldetr} and the ground-truth annotations for the per-class incremental setting on nuScenes in \Cref{fig:nuscenes_ext}.
In the first image, we demonstrate that for some static objects, CL-DETR predicts movement, whereas our method does not.
The second image shows that \net~can detect pedestrians and forecast their movement more reliably.
In the remaining images, we present additional examples showing that our method learns and retains non-linear future trajectories more effectively than CL-DETR.
The latter holds not only for frequently observed objects, such as cars shown in the fourth and fifth rows, but also for other classes, such as bicyclists in the third row and buses in the sixth row.
Our approach also maintains its performance in night scenarios, as shown in the last image.

\begin{figure*}[!ht]
    \centering
    {\includegraphics[width=\linewidth]{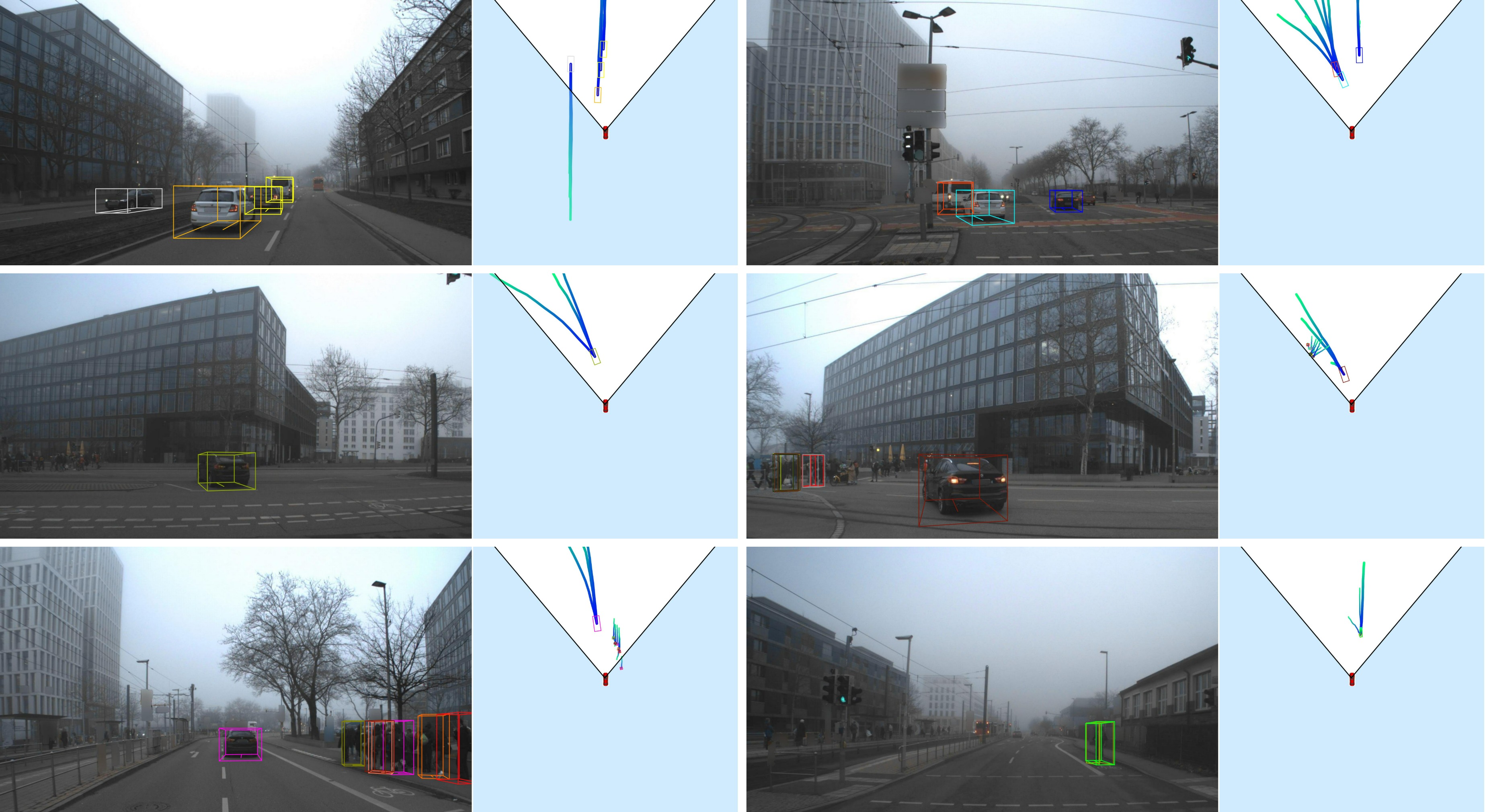}}
    \caption{Extended qualitative zero-shot real-world results recorded with our in-house perception car. The range of the birds-eye-view images is set to \SI{40}{\meter} in the longitudinal and lateral direction of the ego-vehicle, while the prediction horizon is set to \SI{6}{\second}, and the top-$3$ predicted modes per object are shown. The camera's field of view is highlighted in the bird's-eye view projection.}
    \label{fig:realworld_extended}
    \vspace{-0.5cm}
\end{figure*}

\begin{figure}[t]
    \centering
    {\includegraphics[width=0.7\linewidth]{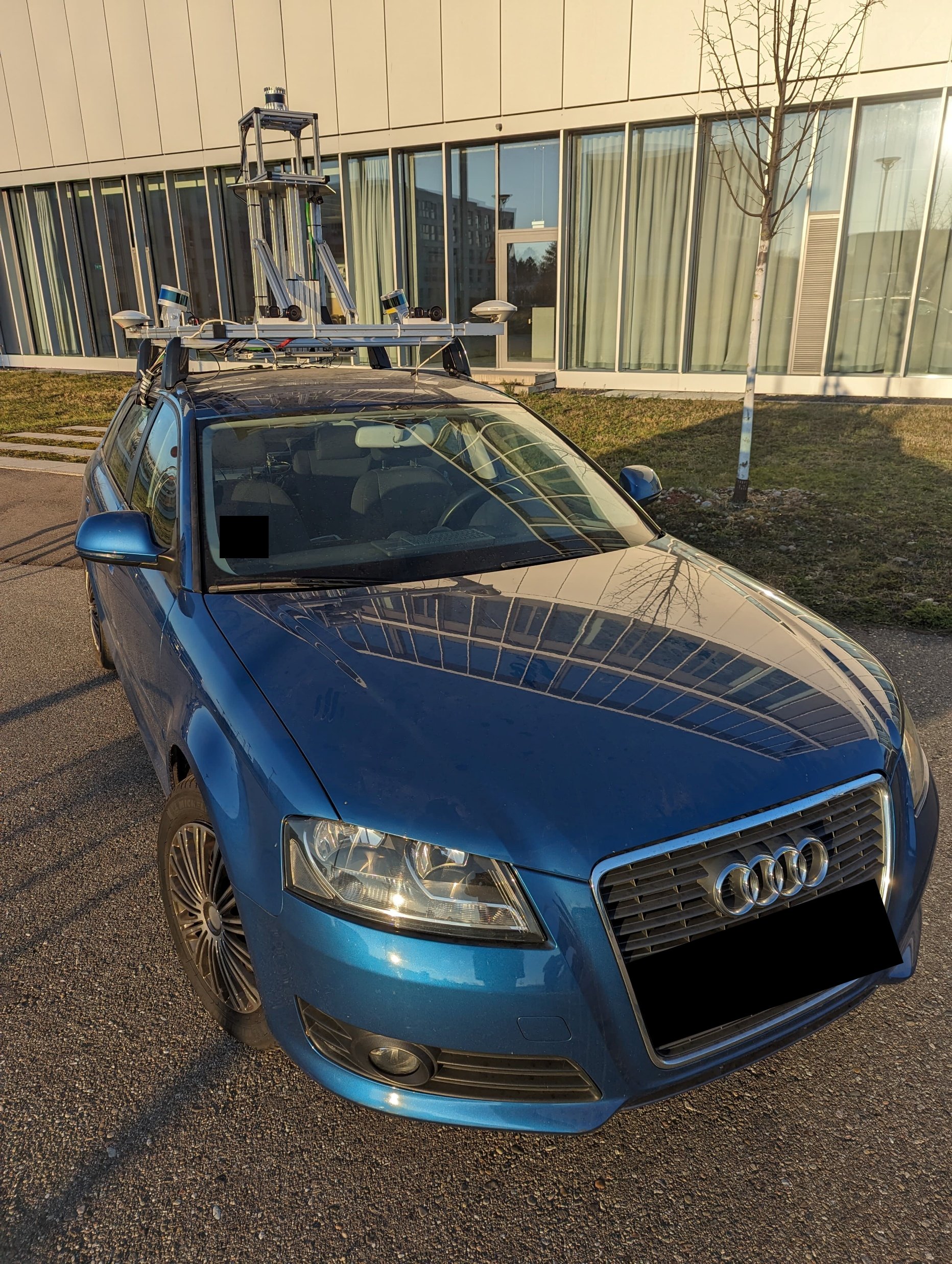}}
    \caption{The self-driving perception car that we used in the real-world experiments.}
    \label{fig:car}
    \vspace{-0.5cm}
\end{figure}

\subsubsection{Planning}
To also visually demonstrate the open-loop planning capabilities of \net, we present results for this additional task in \Cref{fig:planning}.
We demonstrate, using four examples, that the model develops a better understanding of the surrounding agents over the course of training.
In the first two images, \net, after the final incremental step, correctly detects one of the pedestrians and therefore suggests stopping in front of the pedestrian crossing until the pedestrian has crossed the road. In contrast, the model after the first incremental step, which has been trained only on the class car, is not correctly aware of the situation, as it plans to brake too late and to keep standing after the road is clear again.
Similarly, in the third picture, the model after the first step would also not yield to the pedestrians.
We demonstrate that this behavior is not limited to pedestrians: In the last example, the model after the final incremental step notices the bus and plans to stop behind it, whereas the model from the earlier checkpoint does not.
In the first and last example, we observe that the model also behaves similarly when predicting the motion of other traffic participants.
In both cases, the predicted motion of the car behind the ego-vehicle does not account for the presence of pedestrians or the bus.

\subsection{Real-World Experiments}
To demonstrate the zero-shot capability of \net, we evaluate it on sequences from our in-house self-driving perception car shown in \Cref{fig:car}, using only the front-left camera and the global navigation satellite system (GNSS).
The latter is necessary because we need to retrieve the ego-pose of the vehicle to temporally propagate the object queries between frames.
From the recordings, we extract the camera images used for inference at 2 Hz, as in nuScenes, and associate them with the closest GNSS pose, measured at 100 Hz.
The camera images, captured at a resolution of $1200 \times 1920$ (height $\times$ width), are cropped and downsampled to the same image resolution as for the underlying nuScenes training ($256 \times 704$).
It should be noted that there is a significant domain gap between this deployed environment and the nuScenes data due to differences in country, camera, and mounting pose.
For this experiment, we use the per-class incremental setting to train our model only on front-view images from nuScenes.

As shown in \Cref{fig:realworld_extended}, \net~achieves generalizable zero-shot performance when the ego-vehicle is driving.
In the top-left image, our approach correctly estimates the linear motion of oncoming traffic and the vehicles ahead.
\net~also accurately forecasts the turning behavior for two vehicles in the second image.
The next three images show a scene in which the black vehicle ahead first takes a left turn, then stops at a pedestrian crossing to yield to a group of pedestrians before accelerating again.
Our model accurately estimates the vehicle’s turn and predicts a stopping behavior in front of the pedestrians as one of the three most confident modes.
Moreover, it captures the car's subsequent acceleration and reliably predicts pedestrians' movements on the sidewalk.
In the last image, we show that \net~is can detect and forecast the motion of cyclists in our data.

\subsection{Discussion of Limitations}
In the experimental evaluation of this work, we did not enforce that classes may be completely absent in subsequent incremental steps.
This could lead to increased forgetting and, thus, decreased accuracy for these classes, since pseudo-labels cannot be inferred, and our method would rely solely on the variance-based replay mechanism.
Moreover, \net~relies on ground-truth annotations for novel classes.
Our method also treats agents that it was not trained with as background. In the real world, it would be beneficial if the model could still forecast the motion of objects of unknown classes.
\section{Conclusion}
\label{sec:conclusion}
In this paper, we introduced the task of class-incremental motion prediction and proposed the methodological approach \net~to successfully overcome the challenges of this learning setup.
While accounting for storage constraints in autonomous vehicles, we introduced an exemplar replay selection strategy tailored to motion forecasting under an evolving taxonomy of agents, alongside a pseudo-labeling mechanism that uses object detections in future frames and filters false positives by incorporating an open-vocabulary segmentation model.
Extensive experiments and ablations demonstrated that the approach successfully diminishes forgetting while retaining plasticity for learning novel classes, and approaches the performance of upper-bound training with all annotations available simultaneously.
Future work could build on the methodological approach and further enhance it by enabling the prediction of the motion of agents from unknown classes, allowing it to operate in open-world scenarios.

\section{Acknowledgement}
Nikhil Gosala was funded by Qualcomm Technologies Inc., as well as an academic grant from NVIDIA. Nicolas Schischka was funded by the European Union with the HIDDEN project, under grant agreement No 101202228. Views and opinions expressed are however those of the author(s) only and do not necessarily reflect those of the European Union or the European Climate, Infrastructure and Environment Executive Agency (CINEA). Neither the European Union nor the granting authority can be held responsible for them.

\bibliographystyle{IEEEtran}
\bibliography{references}

\end{document}